\documentclass{article}

 \usepackage[main, final]{neurips_2026}

\usepackage[utf8]{inputenc} %
\usepackage[T1]{fontenc}    %
\usepackage{hyperref}       %
\usepackage{url}            %
\usepackage{booktabs}       %
\usepackage{amsfonts}       %
\usepackage{nicefrac}       %
\usepackage{microtype}      %
\usepackage{xcolor}         %

\usepackage{amsmath}
\usepackage{amsthm}
\usepackage{cleveref}
\usepackage{refcount}
\usepackage{enumitem}
\usepackage{amssymb}
\usepackage{tikz}
\usetikzlibrary{shapes}
\usetikzlibrary{patterns.meta}
\usetikzlibrary{calc}
\usetikzlibrary{positioning, fit, arrows.meta, shapes.geometric}
\usepackage{mathtools}
\usepackage{pgfplots}
\usepackage{subcaption}
\usepackage{caption}
\pgfplotsset{compat=1.18}
\usepgfplotslibrary{statistics}
\usepgfplotslibrary{external}
\usepackage{xcolor}
\usepackage{booktabs}
\usepackage{multirow}
\usepackage{wrapfig}
\usepackage{array}
\usepackage{placeins}
 
\definecolor{scp1color}{RGB}{255,225,1}
\definecolor{scp2color}{RGB}{242,170,132}
\definecolor{scp2Bcolor}{RGB}{174,98,100}
\definecolor{scp2Ccolor}{HTML}{916BA4}
\definecolor{scp3color}{RGB}{242,170,132}
\definecolor{scp3Bcolor}{RGB}{174,98,100}
\definecolor{scp3Ccolor}{HTML}{916BA4}
\definecolor{scp4color}{HTML}{916BA4}
\definecolor{scp4Bcolor}{RGB}{174,98,100}
\definecolor{scp4Ccolor}{HTML}{916BA4}
\definecolor{scp5color}{RGB}{242,170,132}
\definecolor{scp6color}{RGB}{242,170,132}
\definecolor{cmcp1color}{RGB}{68,114,196}
\definecolor{nmcp1color}{RGB}{84,130,53}
\definecolor{nmcp2color}{RGB}{156,206,66}
\definecolor{remmcp1color}{RGB}{68,114,196}
\definecolor{relmpp1color}{RGB}{84,130,53}
\definecolor{relmcp2color}{RGB}{156,206,66}

\usepackage{algorithm}
\usepackage[noend]{algpseudocode}
\algblock{Input}{EndInput}
\algnotext{EndInput}
\algblock{Output}{EndOutput}
\algnotext{EndOutput}
\newcommand{\comm}[1]{~~\textit{\footnotesize\textcolor{nmcp1color!80}{\# #1}}}

\newtheorem{theorem}{Theorem}
\newtheorem{assumption}{Assumption}
\newtheorem{corollary}{Corollary}

\newcommand{\ncal}{n_{\mathrm{cal}}}

\newcommand{\neval}{n_{\mathrm{eval}}}
\newcommand{\ns}{n_{\mathrm{s}}}
\newcommand{\nx}{n_{\mathrm{x}}}
\newcommand{\ny}{n_{\mathrm{y}}}
\newcommand{\nq}{n_{\mathrm{q}}}
\newcommand{\nk}{n_{\mathrm{k}}}

\newcommand{\supp}{\mathrm{supp}}
\newcommand{\qe}{q}
\newcommand{\tqe}{\tilde{q}}
\newcommand{\YCP}{\tilde{\mathcal{Y}}}
\newcommand{\Y}{\mathcal{Y}} %
\newcommand{\M}{\mathcal{M}}
\newcommand{\Mtrain}{\mathcal{M}_{\mathrm{train}}}

\newcommand{\Mcal}{\mathcal{M}_{\mathrm{cal}}}
\newcommand{\Mcaltilde}{\tilde{\mathcal{M}}_{\mathrm{cal}}}
\newcommand{\Mout}{\mathcal{R}}%
\newcommand{\Moutk}[1]{\mathcal{R}_{#1}}%
\DeclareMathOperator*{\argmin}{argmin}

\crefname{equation}{}{}
\crefname{figure}{Fig.}{Figures}
\crefname{section}{Sec.}{Sections}
\crefname{table}{Tab.}{Tables}
\crefname{algorithm}{Alg.}{Algorithms}
\crefname{definition}{Def.}{Definitions}
\crefname{problem}{Problem}{Problems}
\crefname{subproblem}{Problem}{Problems}
\crefname{theorem}{Thm.}{Theorems}
\crefname{proposition}{Prop.}{Propositions}
\crefname{corollary}{Corollary}{Corollaries}
\crefname{lemma}{Lem.}{Lemmas}
\crefname{appendix}{Appendix}{}
\crefname{assumption}{Assumption}{Assumptions}

\pgfplotsset{
  mybar/.style={
    width=4.2cm,
    height=3.2cm,
    ybar,
    bar width=0.28cm,
    enlarge x limits=0.18,
    xtick=\empty,          %
    tick align=outside,
    ymajorgrids=true,
    grid style={white, line width=0.6pt},
    axis background/.style={fill=gray!15},
    axis line style={draw=none},
    tick style={draw=none},
    clip=true,
    scaled y ticks=false,
    xmin=0.3, xmax=4.7,
    yticklabel style={font=\small},
    ylabel style={font=\small},
  },
  errbar/.style={
    error bars/y dir=both,
    error bars/y explicit,
    error bars/error bar style={black, line width=0.7pt},
    error bars/error mark options={rotate=90, black, mark size=2pt, line width=0.7pt},
  }
}

\pgfplotsset{
  myplotBase/.style={
    height=5cm,
    xlabel={$e_1$},
    ylabel={$e_2$},
    grid=both,
    grid style={dashed, gray!40},
    tick align=outside,
    axis line style={draw=gray!60},
    tick label style={font=\tiny},
    label style={font=\small},
    title style={font=\small},
    clip=true
  },
  myplotSetVeh/.style={
    myplotBase,
    width=5cm,
    xmin=-4, xmax=4,
    ymin=-7, ymax=7
  },
  myplotSetVeh2/.style={
    myplotBase,
    width=5cm,
    height=5.5cm,
    xmin=-4, xmax=4,
    ymin=-7, ymax=7
  },
  myplotSet/.style={
    myplotBase,
    width=3.7cm,
    xmin=-1, xmax=1,
    ymin=-1, ymax=1
  },
}

\pgfplotsset{
  myboxBase/.style={
    height=3.2cm,
    boxplot/draw direction=y,
    enlarge x limits=0.18,
    xtick=\empty,
    tick align=outside,
    ymajorgrids=true,
    grid style={white, line width=0.6pt},
    axis background/.style={fill=gray!15},
    axis line style={draw=none},
    tick style={draw=none},
    clip=true,
    scaled y ticks=false,
    yticklabel style={font=\footnotesize},
    ylabel style={font=\small},
  },
  myboxVeh/.style={
    myboxBase,
    width=4.8cm,
    xmin=1, xmax=7,
  },
  myboxVeh2/.style={
    myboxVeh, height=4cm,
  },
  mybox/.style={
    myboxBase,
    width=3.7cm, height=4cm,
    xmin=0.8, xmax=4.2,
  },
}

\def\pnone{none}
\newcommand{\sharedlegendVehNew}{%
  \def\legenditems{%
    {scp1color}/{none}/{SCP-1},%
    {scp3color}/{none}/{SCP-2A},%
    {scp4color}/{none}/{SCP-3A},%
    {cmcp1color}/{none}/{RemMCP},%
    {nmcp1color}/{none}/{H-RelMCP-A},%
    {scp3color}/{hatched}/{SCP-2B},%
    {scp4color}/{hatched}/{SCP-3B},%
    {nmcp1color}/{hatched}/{H-RelMCP-B}%
  }%
  \def\boxw{0.3cm}%
  \def\boxh{0.28cm}%
  \def\itemgap{0.2cm}%
  \def\labeloff{0.1cm}%
  \begin{tikzpicture}[baseline, node distance=0pt]
    \coordinate (nextbox) at (0,0);
    \foreach \col/\pat/\lbl [count=\i from 0] in \legenditems {%
      \ifx\pat\pnone
        \node[rectangle, fill=\col,
              minimum width=\boxw, minimum height=\boxh, inner sep=0,
              anchor=west] (box\i) at (nextbox) {};%
      \else
        \node[rectangle,
              pattern={Lines[angle=45, distance=2pt, line width=0.8pt]},
              pattern color=\col,
              minimum width=\boxw, minimum height=\boxh, inner sep=0,
              anchor=west] (box\i) at (nextbox) {};%
      \fi
      \node[font=\footnotesize, anchor=west, inner sep=0]
        (lbl\i) at ($(box\i.east)+(\labeloff,0)$) {\lbl};%
      \coordinate (nextbox) at ($(lbl\i.east)+(\itemgap,0)$);%
    }%
  \end{tikzpicture}%
}

\def\pnone{none}
\newcommand{\sharedlegendVehSingle}{%
  \def\legenditems{%
    {scp1color}/{none}/{SCP-1},%
    {scp3color}/{none}/{SCP-2A},%
    {scp4color}/{none}/{SCP-3A},%
    {cmcp1color}/{none}/{RemMCP},%
    {nmcp1color}/{none}/{H-RelMCP-A}%
  }%
  \def\boxw{0.3cm}%
  \def\boxh{0.28cm}%
  \def\itemgap{0.2cm}%
  \def\labeloff{0.1cm}%
  \begin{tikzpicture}[baseline, node distance=0pt]
    \coordinate (nextbox) at (0,0);
    \foreach \col/\pat/\lbl [count=\i from 0] in \legenditems {%
      \ifx\pat\pnone
        \node[rectangle, fill=\col,
              minimum width=\boxw, minimum height=\boxh, inner sep=0,
              anchor=west] (box\i) at (nextbox) {};%
      \else
        \node[rectangle,
              pattern={Lines[angle=45, distance=2pt, line width=0.8pt]},
              pattern color=\col,
              minimum width=\boxw, minimum height=\boxh, inner sep=0,
              anchor=west] (box\i) at (nextbox) {};%
      \fi
      \node[font=\footnotesize, anchor=west, inner sep=0]
        (lbl\i) at ($(box\i.east)+(\labeloff,0)$) {\lbl};%
      \coordinate (nextbox) at ($(lbl\i.east)+(\itemgap,0)$);%
    }%
  \end{tikzpicture}%
}

\def\psolid{solid}
\newcommand{\sharedlegendSetsVehNew}{%
  \def\legenditems{%
    {scp1color}/{solid}/{SCP-1},%
    {scp3color}/{solid}/{SCP-2A},%
    {scp4color}/{solid}/{SCP-3A},%
    {cmcp1color}/{solid}/{RemMCP},%
    {nmcp1color}/{solid}/{H-RelMCP-A},%
    {scp3color}/{dashed}/{SCP-2B},%
    {scp4color}/{dashed}/{SCP-3B},%
    {nmcp1color}/{dashed}/{H-RelMCP-B}%
  }%
  \def\boxw{0.3cm}%
  \def\boxh{0.28cm}%
  \def\itemgap{0.2cm}%
  \def\labeloff{0.1cm}%
  \begin{tikzpicture}[baseline, node distance=0pt]
    \coordinate (nextbox) at (0,0);
    \foreach \col/\pat/\lbl [count=\i from 0] in \legenditems {%
      \ifx\pat\psolid
        \node[ellipse, line width=1.5pt, draw=\col,
              minimum width=\boxw, minimum height=\boxh, inner sep=0,
              anchor=west] (box\i) at (nextbox) {};%
      \else
        \node[ellipse, line width=1.5pt, dashed, draw=\col,
              minimum width=\boxw, minimum height=\boxh, inner sep=0,
              anchor=west] (box\i) at (nextbox) {};%
      \fi
      \node[font=\footnotesize, anchor=west, inner sep=0]
        (lbl\i) at ($(box\i.east)+(\labeloff,0)$) {\lbl};%
      \coordinate (nextbox) at ($(lbl\i.east)+(\itemgap,0)$);%
    }%
  \end{tikzpicture}%
}

\def\psolid{solid}
\newcommand{\sharedlegendSetsVehSingle}{%
  \def\legenditems{%
    {scp1color}/{solid}/{SCP-1},%
    {scp3color}/{solid}/{SCP-2A},%
    {scp4color}/{solid}/{SCP-3A},%
    {cmcp1color}/{solid}/{RemMCP},%
    {nmcp1color}/{solid}/{H-RelMCP-A}%
  }%
  \def\boxw{0.3cm}%
  \def\boxh{0.28cm}%
  \def\itemgap{0.2cm}%
  \def\labeloff{0.1cm}%
  \begin{tikzpicture}[baseline, node distance=0pt]
    \coordinate (nextbox) at (0,0);
    \foreach \col/\pat/\lbl [count=\i from 0] in \legenditems {%
      \ifx\pat\psolid
        \node[ellipse, line width=1.5pt, draw=\col,
              minimum width=\boxw, minimum height=\boxh, inner sep=0,
              anchor=west] (box\i) at (nextbox) {};%
      \else
        \node[ellipse, line width=1.5pt, dashed, draw=\col,
              minimum width=\boxw, minimum height=\boxh, inner sep=0,
              anchor=west] (box\i) at (nextbox) {};%
      \fi
      \node[font=\footnotesize, anchor=west, inner sep=0]
        (lbl\i) at ($(box\i.east)+(\labeloff,0)$) {\lbl};%
      \coordinate (nextbox) at ($(lbl\i.east)+(\itemgap,0)$);%
    }%
  \end{tikzpicture}%
}
\newcommand{\drawboxlegend}{%
  \def\boxw{0.3cm}\def\boxh{0.28cm}\def\itemgap{0.2cm}\def\labeloff{0.1cm}%
  \begin{tikzpicture}[baseline, node distance=0pt]
    \coordinate (nextbox) at (0,0);
    \foreach \col/\pat/\lbl [count=\i from 0] in \legenditems {%
      \ifx\pat\pnone
        \node[rectangle, fill=\col, minimum width=\boxw, minimum height=\boxh, inner sep=0,
              anchor=west] (box\i) at (nextbox) {};%
      \else
        \node[rectangle, pattern={Lines[angle=45, distance=2pt, line width=0.8pt]}, pattern color=\col,
              minimum width=\boxw, minimum height=\boxh, inner sep=0, anchor=west] (box\i) at (nextbox) {};%
      \fi
      \node[font=\footnotesize, anchor=west, inner sep=0] (lbl\i) at ($(box\i.east)+(\labeloff,0)$) {\lbl};%
      \coordinate (nextbox) at ($(lbl\i.east)+(\itemgap,0)$);%
    }%
  \end{tikzpicture}%
}
\newcommand{\drawsetlegend}{%
  \def\boxw{0.3cm}\def\boxh{0.28cm}\def\itemgap{0.2cm}\def\labeloff{0.1cm}%
  \begin{tikzpicture}[baseline, node distance=0pt]
    \coordinate (nextbox) at (0,0);
    \foreach \col/\pat/\lbl [count=\i from 0] in \legenditems {%
      \ifx\pat\psolid
        \node[ellipse, line width=1.5pt, draw=\col, minimum width=\boxw, minimum height=\boxh,
              inner sep=0, anchor=west] (box\i) at (nextbox) {};%
      \else
        \node[ellipse, line width=1.5pt, dashed, draw=\col, minimum width=\boxw, minimum height=\boxh,
              inner sep=0, anchor=west] (box\i) at (nextbox) {};%
      \fi
      \node[font=\footnotesize, anchor=west, inner sep=0] (lbl\i) at ($(box\i.east)+(\labeloff,0)$) {\lbl};%
      \coordinate (nextbox) at ($(lbl\i.east)+(\itemgap,0)$);%
    }%
  \end{tikzpicture}%
}
\newcommand{\sharedlegendVehNoRel}{%
  \def\legenditems{{scp1color}/{none}/{SCP-1}, {scp3color}/{none}/{SCP-2A}, {scp4color}/{none}/{SCP-3A},
    {cmcp1color}/{none}/{RemMCP}, {scp3color}/{hatched}/{SCP-2B}, {scp4color}/{hatched}/{SCP-3B}}%
  \drawboxlegend}
\newcommand{\sharedlegendVehSingleNoRel}{%
  \def\legenditems{{scp1color}/{none}/{SCP-1}, {scp3color}/{none}/{SCP-2A}, {scp4color}/{none}/{SCP-3A},
    {cmcp1color}/{none}/{RemMCP}}%
  \drawboxlegend}
\newcommand{\sharedlegendSetsVehNoRel}{%
  \def\legenditems{{scp1color}/{solid}/{SCP-1}, {scp3color}/{solid}/{SCP-2A}, {scp4color}/{solid}/{SCP-3A},
    {cmcp1color}/{solid}/{RemMCP}, {scp3color}/{dashed}/{SCP-2B}, {scp4color}/{dashed}/{SCP-3B}}%
  \drawsetlegend}
\newcommand{\sharedlegendSetsVehSingleNoRel}{%
  \def\legenditems{{scp1color}/{solid}/{SCP-1}, {scp3color}/{solid}/{SCP-2A}, {scp4color}/{solid}/{SCP-3A},
    {cmcp1color}/{solid}/{RemMCP}}%
  \drawsetlegend}

\title{Multi-Variable Conformal Prediction: %
Optimizing Prediction Sets using Multiple Calibration Variables}

\author{%
  \textbf{Laura~Lützow$^{1}$} \qquad
  \textbf{Simone Garatti$^{2}$} \qquad
  \textbf{Marco C. Campi$^{3}$} \\[8pt]
  \textbf{Lars Lindemann$^{4}$} \qquad
  \textbf{Matthias Althoff$^{1}$} \\[8pt]
  $^{1}$Technische Universität München \& MCML, Germany \\
  $^{2}$Politecnico di Milano, Italy \\
  $^{3}$Università di Brescia, Italy \\
  $^{4}$Eidgenössische Technische Hochschule Zürich, Switzerland \\[4pt]
  \texttt{laura.luetzow@tum.de} 
}

\begin{document}

\maketitle

\begin{abstract}
Split conformal prediction constructs prediction sets with finite-sample coverage guarantees, but its calibration stage is limited to adjusting a single threshold parameter. Hence, the shape of the prediction sets must be fixed beforehand, typically by fitting it on a reserved part of the calibration data. We introduce multi-variable conformal prediction (MCP), a framework that jointly optimizes and calibrates multiple prediction-set parameters in a single optimization problem, eliminating calibration data split without sacrificing finite-sample coverage guarantees. Building on scenario theory, a principled framework for certifying data-driven decisions, we propose two tractable variants: RemMCP, grounded in constrained optimization with constraint removal, which generalizes split conformal prediction; and RelMCP, based on penalized optimization with constraint relaxation, which also supports non-convex problems, albeit at a higher computational cost. In experiments with ellipsoidal and multi-modal prediction sets, RemMCP and a heuristic, adaptive version of RelMCP consistently meet the target coverage, with lower variance across calibration runs than baselines with calibration data split, and in most settings with smaller or comparable prediction sets. Our code is available at \url{https://github.com/LauraLue/Multi-Variable-Conformal-Prediction}. %

\end{abstract}

\section{Introduction}

\label{sec:intro}
Reliable uncertainty quantification is essential for deploying machine learning models in high-stakes settings, such as scientific inference, autonomous systems, and safety-critical decision-making. Beyond point predictions, these settings demand prediction sets --- regions guaranteed to contain the true output with a user-specified probability --- that are valid under minimal distributional assumptions. Split conformal prediction has emerged as the leading framework for this purpose: given any pre-trained model and a held-out calibration set, it constructs prediction sets with finite-sample marginal coverage guarantees under the mild assumption of exchangeable data~\citep{vovk1999randomness,Lei2013distributionfree,angelopoulos2022gentleintroductionCP}. 

Despite its appeal, split conformal prediction is structurally limited. The calibration stage involves a scalar nonconformity score function and a single scalar nonconformity threshold. This architectural choice forces prediction sets into rigid, often overly conservative shapes that cannot adapt to the geometry of complex outputs or error structures. The straightforward extension to multi-dimensional outputs calibrates a separate threshold per output dimension and corrects the marginal coverage using the Bonferroni method~\citep{stankeviciute2021conformal}, resulting in hyperrectangular prediction sets that are often highly conservative~\citep{feldman2023multiQuantile}. Conservatism can be partially reduced by introducing dimension-specific normalization factors fitted from additional data, which adapt the side lengths of the hyperrectangle to the scale of the residuals~\citep{cleaveland2024conformalLCP}. Other methods reduce conservatism by modeling output dependencies through copulas~\citep{messoudi2021copula,sun2024copula} or graph-based architectures~\citep{cini2025relational}, or by adopting more flexible shapes of prediction sets such as ellipsoids fitted from additional data~\citep{johnstone2021conformal,messoudi2022ellipsoidal,xu2024cpMulti}, zonotopes~\citep{gray2025guaranteed}, norm balls~\citep{braun2025minimumvolumeconformalsets}, multi-modal regions constructed as unions of convex template sets~\citep{tumu2024MultiModalCP}, and nonconvex sets obtained via normalizing flows~\citep{fang2025contra,luo2026densitysortedpredictionsetefficient} or variational autoencoders~\citep{feldman2023multiQuantile}. Other works calibrate directly on vector-valued scores using vine copulas~\citep{park2025semiparametricCP}, optimal transport~\citep{klein2025CPoptimalTransport}, or cell partitions of a multi-dimensional score space~\citep{tawachi2025multidimensionalCP}, or learn multiple set parameters by minimizing an efficiency loss subject to an empirical coverage constraint \citep{bai2022efficientDifferentiableCP}. Yet all of these approaches either provide only approximate coverage guarantees or estimate the shape of the set on data separate from the final calibration step, e.g., through further splitting of the calibration data. The latter reduces data efficiency and inflates the variance across calibration runs. 
Furthermore, most works consider a coverage guarantee that only holds on average over random draws of the calibration data, whereas safety-critical applications require a high-confidence guarantee, i.e., that the coverage exceeds the target with a prescribed probability~\citep{vovk2013condVal}.

In parallel, scenario theory has emerged in control and optimization as a framework for certifying data-driven decisions with finite-sample guarantees~\citep{calafiore2006scenario, garatti2022risk}. While its guarantees are structurally reminiscent of conformal coverage bounds, they arise from a fundamentally different viewpoint based on randomized constraint satisfaction.
Scenario-based methods have been used to construct prediction sets in specific settings, including interval predictors~\citep{campi2009intervalPred,campi2015nonconvex,garatti2019interval,sadeghi2019efficientTraining}, rectangular sets for multivariate data~\citep{deAngelis2021consonantPredictiveBeliefs}, and nonconvex reachable sets in data-driven reachability analysis~\citep{dietrich2024nonconvex}. Recent work has begun to formalize connections between scenario theory and conformal prediction~\citep{coppola2024scenarioCP,lin2024verificationScenarioCP,osullivan2025bridgingconformalpredictionscenario}, which was exploited by zono-conformal prediction~\citep{luetzow2025zonoconformal}, where multiple variables are calibrated simultaneously to construct zonotopic prediction sets. Yet this remains a specific instantiation: a general framework that provides coverage guarantees and admits arbitrary vector-valued score functions and multiple calibration variables has not yet been developed.

\begin{figure}
    \centering
    \definecolor{datasetgray}{RGB}{220, 220, 220}
\definecolor{arrowgray}{RGB}{80, 80, 80}
\begin{tikzpicture}[
    >=Stealth,
    font=\small,
    box/.style={rounded corners=4pt, draw, thick, text width=4.8cm, minimum width=4.8cm, minimum height=0.6cm, align=center, anchor=center},
  every node/.style={},
  arrstyle/.style={->, >=Stealth, line width=0.9pt, color=black},
  dasharr/.style={->, >=Stealth, line width=0.7pt, color=black, dotted},
  labelnode/.style={font=\footnotesize, text centered},
    shapebox/.style={box, fill=scp3color!30, draw=scp3color!70!black},
    shapeboxMCP/.style={box, fill=cmcp1color!30, draw=cmcp1color!70!black},
]

\node[box, fill=datasetgray!40, minimum width=1.2cm, text width=1.2cm, minimum height=2.2cm,
      label={[font=\footnotesize\bfseries, yshift=2pt]above:Calibration data}]
      (dataset) at (0,0) {};

\node[box, fill=datasetgray!80, minimum width=1cm, text width=1cm, minimum height=1cm,text centered, font=\scriptsize]
      (splitA) at (0, 0.5) {Subset 1};

\node[box, fill=datasetgray!80, minimum width=1cm, text width=1cm, minimum height=1cm,text centered, font=\scriptsize]
      (splitB) at (0, -0.5) {Subset 2};

\def\dist{4cm}
\node[shapebox]
     (scpopt)  at ($(splitA.center)+(-\dist,0)$) {\textbf{Optimization} of the prediction set};
     
\node[font=\small\bfseries, color=scp3color, above=0.2cm of scpopt]
(scplabel) {Standard Split Conformal Prediction};

\draw[arrstyle] (splitA.west) -- ++(-0.35,0) |- (scpopt.east);

\node[shapebox]
     (scpcal)  at ($(splitB.center)+(-\dist,0)$) {\textbf{Calibration} of threshold $\tilde{q}^*\in \mathbb{R}$};

\draw[arrstyle] (splitB.west) -- ++(-0.2,0) |- (scpcal.east);

\draw[arrstyle] (scpopt.south) -- (scpcal.north);

\node[shapebox,below=0.4cm of scpcal, text centered] (scppred)
    {\textbf{Prediction set}\\
     $\tilde{\mathcal{Y}}(x,\tilde{q}^*) = \{y \mid \tilde{s}(x,y) \le \tilde{q}^*\}$};

\draw[arrstyle] (scpcal.south) -- (scppred.north);

\node[font=\small\bfseries, color=cmcp1color]
     (mcplabel)  at ($(scplabel.center)+(\dist+\dist,0)$) {Multi-Variable Conformal Prediction};

\node[shapeboxMCP]
     (mcpjoint) at ($(dataset.center)+(\dist,0)$) {\textbf{Joint optimization and calibration} over parameters $q^* \in \mathbb{R}^{\nq}$};

\draw[arrstyle] (dataset.east) -- (mcpjoint.west)
      node[midway, above, font=\scriptsize] {};

\node[shapeboxMCP] (mcppred) at ($(scppred.center)+(\dist+\dist,0)$)
    {\textbf{Prediction set} \\
     $\mathcal{Y}(x,q^*) = \{y \mid s(x,y,q^*) \le \mathbf{0}\}$};

\draw[arrstyle] (mcpjoint.south) -- (mcppred.north);

\begin{scope}[xshift=-1.75cm, yshift=-3.3cm]
\begin{axis}[
    width=5cm,
    height=3cm,
    xlabel={Coverage probability},
    ylabel={},
    xmin=0.92, xmax=1.0,
    ymin=0,
    ytick=\empty,
    xtick=\empty,
    xticklabel style={font=\small},
    xlabel style={font=\small,yshift=0.1cm},
    ylabel style={font=\small},
    axis y line=none,
    axis x line=bottom,
    axis line style={-},
    tick style={color=black, thin},
    clip=false,
]

\addplot[
    color=scp3color,
    line width=1.4pt,restrict x to domain=0.92:inf
] table {figures/fig_helpers/beta_coords_ncal500_eps5_nq1_rho13};

\addplot[
    color=cmcp1color,
    line width=1.2pt,restrict x to domain=0.92:inf
] table {figures/fig_helpers/beta_coords_ncal2000_eps5_nq3_rho25};

\addplot[
    color=black!40!red,
    dashed,dash phase=1pt,
    line width=1.0pt,
] coordinates {(0.95, 0) (0.95, 96)};
\node[
    anchor=east,
    font=\scriptsize,
    text=black!30!red,align=right,inner xsep=2pt
] at (axis cs: 0.95, 62) {Target\\coverage\\$1-\varepsilon$};
\node[font=\footnotesize\bfseries, scp3color, name=highvarnode, align=center, anchor=east] at (axis cs: 0.92, 18) {Conservative,\\ high variance};
\node[font=\footnotesize\bfseries, cmcp1color, name=lowvarnode, align=center, anchor=west] at (axis cs: 1, 18) {Tight,\\ low variance};
\end{axis}
\end{scope}

\draw[dasharr, color=scp3color!70!black] 
    (scppred.south) |- (highvarnode.west);
\draw[dasharr, color=cmcp1color!70!black] 
    (mcppred.south) |- (lowvarnode.east);

\end{tikzpicture}
    \caption{To optimize the prediction-set shape, split conformal prediction (SCP) reserves part of the calibration data and calibrates a single threshold on the rest, whereas multi-variable conformal prediction (MCP) jointly optimizes and calibrates multiple shape parameters on all calibration data. Both can certify the high-confidence coverage guarantee $\mathbb{P}(\eta\geq 1-\varepsilon)\geq 1-\beta$, but MCP yields a narrower and less conservative coverage distribution. %
    }
    \label{fig:overview}
\end{figure}

We close this gap with multi-variable conformal prediction (MCP), a framework that extends the conformal calibration stage to accommodate vector-valued score functions and multiple simultaneous optimization variables. Unlike existing approaches, MCP couples prediction set design and calibration into a single optimization problem, as illustrated in \cref{fig:overview} --- allowing the shape of sets to adapt directly to the available data and enabling joint calibration of multiple parameters without sacrificing distribution-free coverage guarantees or requiring an additional split of the calibration data. 
Our main contributions are:
\begin{itemize}[leftmargin=*, nosep]
    \item We introduce MCP, a general framework that extends conformal calibration to vector-valued score functions and multiple simultaneous optimization variables, eliminating the need to split the calibration data for optimizing the shape of prediction sets (\cref{sec:multiVarCP}).
    \item We develop RemMCP (\cref{subsec:RemMCP}), a variant based on scenario optimization with constraint removal that recovers split conformal prediction as a special case and admits finite-sample expected-coverage and high-confidence coverage guarantees.
    \item We develop RelMCP (\cref{subsec:RelMCP}), a variant based on unconstrained optimization with constraint relaxation that supports non-convex score functions and provides a high-confidence coverage guarantee for a predetermined grid of penalty parameters; its adaptive, heuristic version H-RelMCP is not covered by this guarantee but meets the target empirically.
    \item We provide guidance on the design of score and cost functions to trade off expressiveness, computational complexity, and conservatism (\cref{subsec:scoreCostDesign}).
    \item Through numerical experiments, we show that MCP consistently meets the target coverage, with lower variance across calibration runs than split conformal prediction baselines with calibration data split, while often yielding smaller prediction sets (\cref{sec:experiments}).
\end{itemize}
Together, these results establish MCP as a flexible and theoretically grounded extension of conformal inference, bridging ideas from split conformal prediction and scenario-based optimization.

\section{Preliminaries}
\label{sec:prel}

This section introduces the two methodological pillars of the proposed framework. We first review scenario theory and its finite-sample guarantees for data-driven optimization problems, which form the theoretical backbone of our approach. We then revisit split conformal prediction, presenting it in a formulation that extends naturally to the multi-variable setting developed in the subsequent section.

\subsection{Scenario Theory for Data-Driven Optimization}\label{subsec:scenario}

Scenario theory provides a principled framework for optimization under uncertainty.  Originally introduced in \cite{calafiore2005uncertain,calafiore2006scenario,campi2008exact} as an algorithmic tool for robust design, it has since evolved into a comprehensive methodology for data-driven decision-making; see \cite{campi2021scenario} for a recent survey. We focus on two schemes most relevant to MCP: optimization with constraint removal and optimization with constraint relaxation. Let $\qe \in \mathbb{R}^{\nq}$ denote the decision variable and let $\Mcal = \{\delta^{(m)}\}_{m=1}^{\ncal} \subset \Delta^{\ncal}$ be independent and identically distributed (i.i.d.) samples drawn from an unknown distribution over the sample space $\Delta$. Each sample $\delta$ induces a constraint $\qe \in \mathcal{Q}_{\delta}$, while preferences over feasible solutions are encoded through a cost function $J(\qe)$. Since the optimizer $\qe^*_{\Mcal}$ depends on random samples, it is itself a random variable whose reliability is characterized by the {violation probability}
$V(\qe^*_{\Mcal})= \mathbb{P}\left(\delta \in \Delta \colon \qe^*_{\Mcal} \not\in \mathcal{Q}_{\delta}  \mid \Mcal\right)$, which is the probability that a previously unseen sample violates the learned constraint. 

\textbf{Optimization with constraint removal.}
To balance solution quality and conservatism, we allow the optimizer to discard a subset $\Mout \subset \Mcal$ of sampled constraints, yielding the scenario program
\begin{align}\label{eq:SP}
\qe^*_{\Mcal}
= \argmin_{\qe \in \mathbb{R}^{\nq}}~ J(\qe)
\quad
\text{s.t.}~~ 
\qe \in \mathcal{Q}_{\delta},
~~ \forall \delta^{(m)}\in \Mcal \setminus \Mout.
\end{align}
This class of schemes is studied in~\citep{campi2011samplingDiscard}. Among the many admissible removal schemes, we adopt the one from~\citep{romao2023probabilistic,romao2023exactFeasibility}, which is especially well-suited to MCP and admits particularly tight guarantees.
Let $\supp(\qe^*_{\Mcal})$ denote the {support set} of the optimizer, which contains the samples $\delta \in \Mcal$ whose removal would change $\qe^*_{\Mcal}$. Given a removal budget $\rho \in \mathbb{N}_0$, the set $\Mout = \Moutk{\rho}$ is constructed iteratively via~\citep{romao2023probabilistic,romao2023exactFeasibility}
\begin{align}\label{eq:Moutk}
    \Moutk{k}= \begin{cases}
        \emptyset & \text{if }k=0, \\
        \Moutk{k-1} \cup \supp(\qe_{k-1}^*) \cup \mathcal{Z}_{k-1}& \text{otherwise},
\end{cases}
\end{align}
where $\qe_k^*$ is the optimizer at stage $k$ for $\Mcal \setminus \Moutk{k}$, and $\mathcal{Z}_{k-1}$ consists of $\nq - |\supp(\qe_{k-1}^*)|$ additional scenarios from $\Mcal \setminus (\Moutk{k-1} \cup \supp(\qe_{k-1}^*))$, selected via a lexicographic tie-breaking rule to ensure $|\Moutk{k}| = k \nq$ at each stage. 
The guarantees for this scheme rely on three assumptions. 

\begin{assumption}\label{assumption:unique}
For any $\ncal \in \mathbb{N}$ and $\Mcal \subset \Delta^{\ncal}$, the optimal solution of \cref{eq:SP} exists and is unique.
\end{assumption}

\begin{assumption}\label{assumption:nondenerate}
For each $k = 0, \dots, \rho$, the scenario program in \cref{eq:SP} over $\Mcal \setminus \Moutk{k}$ is nondegenerate with probability one: solving \cref{eq:SP} using only $\supp(\qe^*_k)$ yields the same solution as using $\Mcal\setminus \Moutk{k}$.
\end{assumption}

\begin{assumption}\label{assumption:convex}
The objective $J$ is convex in $\qe$ and $\mathcal{Q}_{\delta}$ is a convex set for all $\delta \in \Delta$.
\end{assumption}

Uniqueness in \cref{assumption:unique} can be relaxed via a convex tie-breaking rule. Existence and convexity (\cref{assumption:unique,assumption:convex}) can be enforced by construction through appropriate choices of the cost function and constraint sets. Under convexity, the nondegeneracy condition in \cref{assumption:nondenerate} is mild: it implies that the constraints are in general position and have no accumulation points~\citep{romao2023exactFeasibility}. In practice, these assumptions are straightforward to satisfy. 

Under \cref{assumption:unique,assumption:convex,assumption:nondenerate}, the optimizer $\qe_{\rho}^*$ 
satisfies~\citep[Thm.~4]{romao2023exactFeasibility}
\begin{align}\label{eq:varCovScenario}
\mathbb{P}\left(V(\qe_{\rho}^*)\leq \varepsilon\right)
\geq 1- 
\sum_{j=0}^{\nq(\rho+1)-1}
\binom{\ncal}{j}\varepsilon^j(1-\varepsilon)^{\ncal-j},
\end{align}
bounding the confidence with which the violation probability does not exceed $\varepsilon$. The expected violation probability satisfies the complementary bound~\citep[Thm.~3]{osullivan2025bridgingconformalpredictionscenario}
\begin{align}\label{eq:expCovScenario}
\mathbb{E}\left[V(\qe_{\rho}^*)\right] \leq \frac{\nq(\rho+1)}{\ncal+1},
\end{align}
which shows that the average violation probability over random draws of $\Mcal$ decreases as the calibration set grows or the removal budget shrinks. Notably, \cref{eq:expCovScenario} continues to hold when the samples in $\Mcal$ are not i.i.d. but merely exchangeable, since the bound follows from the invariance of the distribution to permutations of the observations. %
The following two additional assumptions are needed for the bound in \cref{eq:varCovScenario} to hold with equality~\citep[Thm.~5]{romao2023exactFeasibility}:

\begin{assumption}\label{assumption:fullSupport}
For each $k = 0, \dots, \rho$, the scenario program in \cref{eq:SP} over $\Mcal \setminus \Moutk{k}$ is fully supported with probability one: for any $\Mcal \subset \Delta^{\ncal}$ with $\ncal > \nq$, the support set $\supp(\qe^*_{\Mcal})$ has cardinality $\nq$ with probability one.
\end{assumption}

\begin{assumption}\label{assumption:scenarioViolation}
For any $\Mcal \subset \Delta^{\ncal}$ and $\Mcaltilde \subset \Mcal$, for any $k = 0, \dots, \rho$ and $\delta \in \supp(\tilde{q}_k)$, we have $q^*(\M) \notin \mathcal{Q}_{\delta}$ for any $\M \subset \Mcaltilde \setminus (\bigcup_{j=0}^{k-1} \supp(\tilde{q}_j) \cup \{\delta\})$ with $|\M| = \nq$, where $\tilde{q}_k$ is the optimizer at stage $k$ for $\Mcaltilde$.
\end{assumption}
The violation condition (\cref{assumption:scenarioViolation}) requires that each support sample is violated by every solution that can be obtained from $\nq$ of the samples remaining after it and the earlier support sets are removed.

\textbf{Optimization with constraint relaxation.}
As an alternative to constraint removal, constraint relaxation~\citep[Sec.~4.1]{garatti2024nonconvex} replaces hard constraints with penalized violations. This approach is also applicable to non-convex problems and can, in some cases, yield less conservative solutions, albeit with coverage guarantees that are evaluated a posteriori rather than specified a priori. We adopt this approach in \cref{subsec:RelMCP} to derive the RelMCP formulation.

\subsection{Split Conformal Prediction} \label{subsec:splitCP}
Split conformal prediction (SCP), also known as inductive conformal prediction, is a distribution-free method for constructing prediction sets with finite-sample marginal coverage guarantees~\citep{angelopoulos2022gentleintroductionCP}. 
Let
$\M$ 
be a dataset consisting of samples $(x,y)$ from an unknown distribution over the data space $\Delta$, which we randomly split into a training dataset $\Mtrain$ and a calibration dataset $\Mcal= \{(x^{(m)}, y^{(m)})\}_{m=1}^{\ncal} \subset \Delta^{\ncal}$. 
SCP is based on the following procedure:
\begin{enumerate}[leftmargin=*, nosep]
    \item \textbf{Training:} 
    \begin{enumerate}
        \item We fit a point predictor $f\colon\mathbb{R}^{\nx} \to \mathbb{R}^{\ny}$ using $\Mtrain$. 
        \item We define a score function $\tilde{s} \colon\mathbb{R}^{\nx} \times \mathbb{R}^{\ny} \to \mathbb{R}$, which measures how poorly a candidate output $y$ conforms to the prediction $f(x)$. The structure of the score function determines the geometry of the prediction set, while the scale of the prediction sets is calibrated in the subsequent step to achieve the desired coverage. Optionally, the score function may be fitted using additional data~\citep{johnstone2021conformal,feldman2023multiQuantile,tumu2024MultiModalCP}.
    \end{enumerate}  
    \item \textbf{Calibration:} 
    \begin{enumerate}
        \item We compute the outlier budget for a target miscoverage level $\alpha \in (0,1)$ as 
        \begin{align}\label{eq:noutCP}
            \rho =\lfloor \alpha(\ncal+1) \rfloor -1. 
        \end{align}
        \item We set the conformal threshold $\tqe^*\in\mathbb{R}$ to the $(\ncal-\rho)$-smallest nonconformity score in $\{\tilde{s}(x^{(m)}, y^{(m)})\}_{m=1}^{\ncal}$, 
        which is obtained by applying $\tilde{s}$ to the calibration set $\Mcal$. This is equivalent to setting $\tqe^*$ to the empirical $(1 - \alpha)$-quantile of the scores augmented by $\infty$ \citep{tumu2024MultiModalCP}.
    \end{enumerate}   
    \item \textbf{Prediction:} The prediction set for a new input $x$ is constructed via 
    \begin{align}
        \YCP(x, \tqe^*) = \left\{ y \in \mathbb{R}^{\ny} \mid \tilde{s}(x, y) \leq \tqe^* \right\}. \label{eq:YCP}
    \end{align}
\end{enumerate}
Let $\tilde{\eta}(\tqe^*)=\mathbb{P}\bigl((x,y)\in\Delta\colon y\in\YCP(x,\tqe^*) \mid \Mcal \bigr)$ denote the coverage probability of the prediction set for a new test point $(x,y)$.
Under exchangeability of the data points from $\Mcal$ and the test point, this construction guarantees finite-sample coverage~\citep[Sec.~4.1]{angelopoulos2026theoreticalfoundationsCP}
\begin{align}\label{eq:YCP_cov}
\mathbb{E} \left[ \tilde{\eta}(\tqe^*) \right] 
\geq 1 - \alpha,
\end{align}
where the expectation is over the calibration data.
If the data points satisfy the stronger condition of being i.i.d. and if the score distribution is continuous, the random variable $\tilde{\eta}(\tqe^*)$ follows a Beta distribution~\citep[Sec.~4.1]{angelopoulos2026theoreticalfoundationsCP}:
\begin{align} \label{eq:YCP_beta}
\tilde{\eta}(\tqe^*) \sim \mathrm{Beta}(\ncal - \rho, \rho+1),
\end{align}
which allows computing the confidence level $1-\beta$ in the high-confidence coverage guarantee $\mathbb{P}(\tilde{\eta}(\tqe^*) \geq 1 - \varepsilon)\geq 1-\beta$ for any miscoverage level $\varepsilon\in(0,1)$. %

In regression settings, a standard choice of nonconformity score is the $p$-norm of the residual,
$\tilde{s}(x,y) = \|y - f(x)\|_{p}$, which induces prediction regions given by $p$-norm balls centered at $f(x)$ \citep{braun2025minimumvolumeconformalsets}. 
Alternatively, one may employ dimension-wise scores 
\begin{align}
\tilde{s}_{j}(x,y) = \bigl|y_{(j)}-f_{(j)}(x)\bigr|, 
\label{eq:CPdimScore}
\end{align}
and calibrate a separate threshold for each dimension $j = 1,\dots,\ny$. 
This yields axis-aligned rectangular prediction sets. 
Joint coverage can be ensured via a Bonferroni correction by calibrating each marginal model at level $1 - \alpha / \ny$ \citep{stankeviciute2021conformal}. 
Multi-modal prediction regions can be constructed using 
$\tilde{s}(x,y) = \min_{j} \left(\tilde{s}_{\theta_j}(x,y) \right)$,
where $\tilde{s}_{\theta_j}$ are normalized score functions parameterized by $\theta_j$ and fitted on additional data \citep{tumu2024MultiModalCP}. 
For example, unions of ellipsoids arise by defining
$\tilde{s}_{\theta_j}(x,y) 
= (y-f(x) - \hat{c}_j)^\top \hat{\Sigma}^{-1}_j (y-f(x) - \hat{c}_j)$, with fitted parameters $\theta_j = (\hat{\Sigma}_j,\; \hat{c}_j)$, $\hat{\Sigma}_j \in \mathbb{R}^{\ny \times \ny}$, $\hat{\Sigma}_j \succ 0$, $\hat{c}_j \in \mathbb{R}^{\ny}$.

\section{Multi-Variable Generalization}
\label{sec:multiVarCP}

Classical SCP relies on a scalar score function $\tilde{s}$ and calibrates a single scalar threshold $\tqe^*$. We extend this framework by introducing a vector-valued score function $s(x, y, \qe) \in \mathbb{R}^{\ns}$ and a parameter vector $\qe \in \mathbb{R}^{\nq}$ that determine whether an output $y$ conforms to the prediction $f(x)$ of a point predictor $f$. Given $\Mcal = \{(x^{(m)}, y^{(m)})\}_{m=1}^{\ncal} \subset \Delta^{\ncal}$, the goal is to find $\qe^*$ such that the prediction set
\begin{align}\label{eq:YMCP}
\Y(x,\qe^*)=\{y\in\mathbb{R}^{\ny}\mid s(x,y,\qe^*)\leq \mathbf{0} \}
\end{align}
satisfies a prescribed coverage guarantee on $\eta(\qe^*)=\mathbb{P}\left((x,y)\in\Delta\colon y\in\Y(x,\qe^*) \mid \Mcal \right)$ while optimizing a user-defined objective $J(\qe)$ that encodes preferences, such as minimal prediction set volume. 
We consider two types of coverage guarantees. For a target miscoverage level $\alpha\in(0,1)$, the expected coverage guarantee is
\begin{align} \label{eq:expGuarantee}
    \mathbb{E}[\eta(\qe^*)]\geq 1-\alpha
\end{align}
where the expectation is over the calibration data. For a target miscoverage level $\varepsilon\in(0,1)$ and confidence level $1-\beta$ with $\beta\in(0,1)$, the high-confidence guarantee is
\begin{align} \label{eq:confGuarantee}
    \mathbb{P}\left(\eta(\qe^*)\geq 1-\varepsilon\right)\geq 1-\beta.
\end{align}
As in SCP, the expected guarantee in \cref{eq:expGuarantee} only requires the calibration data and the test point to be exchangeable, whereas the high-confidence guarantee in \cref{eq:confGuarantee} requires them to be i.i.d.

We develop two tractable formulations for multi-variable conformal prediction (MCP). The first is based on a constrained scenario program with constraint removal, which recovers SCP as a special case but requires the score function to induce convex sets of feasible parameters. The second relies on unconstrained optimization with constraint relaxation; this approach applies more broadly, including non-convex settings. A comparison of both formulations is provided in Appendix~\ref{sec:comparison}.
Design considerations for the score and cost function are discussed in Section~\ref{subsec:scoreCostDesign}.

\subsection{MCP via Constraint Removal} \label{subsec:RemMCP}
MCP based on constrained optimization with constraint removal (RemMCP) constructs prediction sets that satisfy either the expected coverage guarantee in \cref{eq:expGuarantee} or the high-confidence guarantee in \cref{eq:confGuarantee}, depending on the choice of the outlier budget. It proceeds as follows:
\begin{enumerate}[leftmargin=*, nosep]
    \item \textbf{Training:}
    \begin{enumerate}
        \item We fit a point predictor $f\colon\mathbb{R}^{\nx} \to \mathbb{R}^{\ny}$ using $\Mtrain$. 
        \item We define a score function $s(x,y,\qe)\in\mathbb{R}^{\ns}$ such that the set $\{\qe \mid s(x,y,\qe)\leq\mathbf{0}\}$ is convex for all $(x,y)$, e.g., because $s$ is convex in $\qe$, and a convex cost function $J(\qe)\in\mathbb{R}$. Both can optionally be fitted using additional data that is independent of $\Mcal$.        
    \end{enumerate}
    \item \textbf{Calibration:}
    \begin{enumerate}    
    \item We compute the outlier budget $\rho$ according to the targeted guarantee:
    \begin{itemize}[align=left, labelindent=0pt, labelsep=0.4em, nosep]
        \item \emph{Expected coverage} \cref{eq:expGuarantee} for a target miscoverage level $\alpha$: %
        \begin{align} \label{eq:noutMCP}
            \rho = \Bigl\lfloor \frac{\alpha(\ncal+1)}{\nq} \Bigr\rfloor -1.
        \end{align}
        \item \emph{High-confidence coverage} \cref{eq:confGuarantee} for a target miscoverage level $\varepsilon$ and confidence $1-\beta$: %
        \begin{align}\label{eq:noutMCPconf}
            \rho = \max\bigl\{\rho'\in\mathbb{N}_0 \mid \bar{\beta}(\rho')\leq\beta\bigr\},\ \text{with}\ 
            \bar{\beta}(\rho) = \smashoperator[l]{\sum_{j=0}^{\nq(\rho+1)-1}}
            \binom{\ncal}{j}\varepsilon^j(1-\varepsilon)^{\ncal-j}.
        \end{align}
    \end{itemize}
    If the right-hand side of \cref{eq:noutMCP} is smaller than 0 or $\bar{\beta}(0)>\beta$ in \cref{eq:noutMCPconf}, no feasible $\rho\in\mathbb{N}_0$ exists. In this case, we need to increase the size of $\Mcal$ or, alternatively, reduce the parameter dimension $\nq$, which requires redefining the score and cost functions.
    \item We compute the parameter $\qe^* = \qe_{\rho}^*$ by solving the cascading optimization problems
    \begin{align}\label{eq:MCPprogram}
    \qe_k^* = \argmin_{\qe\in\mathbb{R}^{\nq}}~ J(\qe) \quad 
    \text{s.t.}~~  s(x^{(m)},y^{(m)},\qe)\leq \mathbf{0},
    ~~ \forall (x^{(m)},y^{(m)})\in \Mcal\setminus\Moutk{k}, 
    \end{align}
    for $k=0,\dots,\rho$, with $\Moutk{k}$ constructed via \cref{eq:Moutk}.
\end{enumerate}
    \item \textbf{Prediction:} The prediction set for a new input $x$ is constructed via \cref{eq:YMCP}.
\end{enumerate}

This prediction set construction guarantees that every calibration data point $y^{(m)}$ in $\Mcal\setminus\Moutk{\rho}$ is covered by its respective prediction set $\Y(x^{(m)},\qe^*)$.
We obtain the following coverage guarantees:

\begin{theorem}[Probabilistic Guarantees of RemMCP]\label{thm:multiVarCP}
    Let \cref{assumption:unique,assumption:convex,assumption:nondenerate} hold for the optimization problems in \cref{eq:MCPprogram}. Then, for any $\rho\in\mathbb{N}_0$ with $\nq(\rho+1)<\ncal$, the following holds:
    \begin{enumerate}[label=(\alph*), leftmargin=*, nosep]
        \item \label{thm:multiVarCP_exp} \emph{Expected coverage.} If the calibration samples in $\Mcal$ and the test point are exchangeable, RemMCP satisfies the expected coverage guarantee in \cref{eq:expGuarantee} with $\alpha = \nq(\rho+1)/(\ncal+1)$; in particular, choosing $\rho$ via \cref{eq:noutMCP} ensures \cref{eq:expGuarantee} for a user-defined target $\alpha$.
        \item \label{thm:multiVarCP_conf} \emph{High-confidence coverage.} If the calibration samples in $\Mcal$ and the test point are i.i.d., RemMCP satisfies the high-confidence guarantee in \cref{eq:confGuarantee} with $\beta=\bar{\beta}(\rho)$ from \cref{eq:noutMCPconf}; in particular, choosing $\rho$ via \cref{eq:noutMCPconf} ensures \cref{eq:confGuarantee} for a user-defined target $\beta$.
        \item \label{thm:multiVarCP_exact} \emph{Exact coverage distribution.} If the calibration samples in $\Mcal$ and the test point are i.i.d.\ and, additionally, \cref{assumption:fullSupport,assumption:scenarioViolation} hold, the coverage follows exactly
        \begin{align}\label{eq:betaMCP}
            \eta(\qe^*)\sim\mathrm{Beta}\bigl(\ncal-\nq(\rho+1)+1, \nq(\rho+1)\bigr).
        \end{align} 
        In this case, $\bar{\beta}(\rho)=I_{1-\varepsilon}\bigl(\ncal-\nq(\rho+1)+1,\,\nq(\rho+1)\bigr)$ is the cumulative distribution function of \cref{eq:betaMCP} evaluated at $1-\varepsilon$, so that \cref{eq:confGuarantee} holds with equality for $\beta=\bar{\beta}(\rho)$.
    \end{enumerate}
\end{theorem}
We next show that SCP is a special case of RemMCP and that RemMCP is less conservative than dimension-wise SCP. All proofs are provided in Appendix~\ref{sec:proofs}.

\begin{corollary}\label{cor:RemMCP2scp}
    The prediction sets \cref{eq:YCP} and coverage guarantees \cref{eq:YCP_cov,eq:YCP_beta} of SCP with scalar score $\tilde{s}(x,y)$ and continuous score distribution are recovered from RemMCP with $\rho$ from \cref{eq:noutMCP} by setting
    \begin{subequations} \label{eq:MCP2CP}
    \begin{align} 
            s(x,y,\qe) &= \tilde{s}(x,y) - \qe, \label{eq:MCP2CPscore} \\
            J(\qe) &= \qe.\label{eq:MCP2CPcost}
    \end{align}        
    \end{subequations}    
\end{corollary}

\begin{corollary}\label{cor:dSCP}
The prediction sets obtained by RemMCP with $\rho$ from \cref{eq:noutMCP} and
\begin{subequations}\label{eq:MCPdSCP}
\begin{align}
    s(x,y,\qe)&= \begin{bmatrix}
        -y + f(x) - \qe \\
        \hphantom{-}y-f(x)-\qe 
    \end{bmatrix}, \label{eq:MCPdSCPscore}\\
    J(\qe) &= \mathbf{1}^\top \qe, \label{eq:MCPdSCPcost}
\end{align}    
\end{subequations}
assuming a continuous score distribution,  
are contained in the prediction sets produced by the dimension-wise SCP method in \cref{eq:CPdimScore} with Bonferroni correction~\citep{stankeviciute2021conformal} for any target miscoverage level $\alpha\in(0,1)$.
\end{corollary}

Since \cref{cor:dSCP} holds for every input $x$ with probability one over the calibration data, it is stronger than an average or high-probability volume guarantee, and the containment can be strict when samples are extreme in several dimensions, as dimension-wise SCP then discards such samples repeatedly, whereas RemMCP always discards $\rho\ny$ distinct samples.

\subsection{MCP via Constraint Relaxation} \label{subsec:RelMCP}

An alternative approach for MCP, which we name RelMCP, bases the calibration phase on unconstrained optimization via constraint relaxation while targeting the high-confidence guarantee in \cref{eq:confGuarantee} for a prescribed confidence level $1-\beta$. RelMCP proceeds as follows.
\begin{enumerate}[leftmargin=*, nosep]
    \item \textbf{Training:} As in \cref{subsec:RemMCP}, but the score and cost function do not need to be convex in $\qe$.
    \item \textbf{Calibration:} For a grid of penalty parameters $\phi_1,\dots,\phi_{\neval}>0$, which control the trade-off between objective optimality and empirical constraint satisfaction, we solve
    \begin{align}\label{eq:RelMCPprogram}
    \qe_i^*=\argmin_{\qe\in\mathbb{R}^{\nq}} \; J(\qe)+\phi_i\sum_{m=1}^{\ncal}\xi_m(\qe),
    \end{align}
    with $\xi_m(\qe)=\max\left\{0,\max_{j\in\{1,\dots,\ns\}} s_{(j)}(x^{(m)},y^{(m)},\qe)\right\}$ and $i=1,\dots,\neval$.
    The achievable miscoverage $\varepsilon_i$ for each $\qe_i^*$ is evaluated a posteriori as the solution to
    \begin{align}\label{eq:nonconvexCovTradeoff}
    \frac{\beta}{\neval\ncal}\sum_{j=d_i}^{\ncal-1}
    \frac{\binom{j}{d_i}}{\binom{\ncal}{d_i}}
    \frac{1}{(1-\varepsilon_i)^{\ncal-j}}=1,
    \end{align}
    where $d_i$ is the solution complexity~\citep{garatti2024nonconvex}; namely, the number of samples satisfying $\xi_m(\qe_i^*)>0$, plus the cardinality of the smallest additional subset of samples that, together with them, uniquely determines $\qe_i^*$. 
    Among the solutions $\qe_i^*$ that achieve the target coverage, i.e., with $\varepsilon_i\leq \varepsilon$, we select the one with the smallest penalty $\phi_i$ as $\qe^*$.   
    \item \textbf{Prediction:} The prediction set for a new input $x$ is constructed via \cref{eq:YMCP}. %
\end{enumerate}

\begin{theorem}[Probabilistic Guarantees of RelMCP]\label{thm:multiVarCP_nonconvex}
    Let the calibration samples in $\Mcal$ and the test point be i.i.d., let the penalty parameters $\phi_1,\dots,\phi_{\neval}$ be fixed before calibration, and let $\varepsilon_i$ in \cref{eq:nonconvexCovTradeoff} be computed with the solution complexity $d_i$ or an upper bound thereof.
    Then RelMCP satisfies the high-confidence bound in \cref{eq:confGuarantee}.
\end{theorem}

A larger grid lowers the certifiable coverage $1-\varepsilon_i$ in \cref{eq:nonconvexCovTradeoff}, since $\beta$ is shared among all $\neval$ penalties. As a suitable grid is hard to predefine and the exact solution complexity is hard to compute, we also propose Heuristic RelMCP (H-RelMCP), which selects the penalties adaptively and approximates $d_i$ heuristically as shown in \cref{alg:RelMCP} (see Appendix~\ref{sec:RelMCPalg} for a discussion). H-RelMCP is an empirically motivated extension not covered by \cref{thm:multiVarCP_nonconvex}; in our experiments, however, its share of runs that missed the target coverage remained far below the chosen $\beta$ 
(Appendix~\ref{subsec:varConfidence}).

\begin{algorithm}[h!]
\caption{H-RelMCP Calibration}\label{alg:RelMCP}
\begin{algorithmic}[1]

\Statex \textbf{Inputs:} Calibration data $\mathcal{M}_\mathrm{cal}$, score function $s(x, y, q)$, cost function $J(q)$, target miscoverage $\varepsilon \in (0,1)$, target misconfidence $\beta \in (0,1)$, initial penalty weight $\phi_1 > 0$, increase factor $f_+ > 1$, decrease factor $f_- \in (0,1)$, soft iteration limit $i_\mathrm{soft} \in \mathbb{N}$, maximum number of iterations $i_\mathrm{max} \geq i_\mathrm{soft}$

\vspace{0.2cm}

\State $i \leftarrow 1$;\enspace $i_\mathrm{val} \leftarrow 0$;\enspace $i_\mathrm{inv} \leftarrow 0$ \comm{Initialize iteration counters}

\While{$i \leq i_\mathrm{max}$}
  \State $\qe_i^* \leftarrow$ solve \cref{eq:RelMCPprogram} with penalty $\phi_i$ ~~\comm{Solve optimization problem}
  \State $d_i \leftarrow v_i + \min(\nq,\, \ncal - v_i)$ with $v_i = |\{m \colon \xi_m(\qe_i^*) > 0\}|$ ~\comm{Heuristic solution complexity}
  \State $\varepsilon_i \leftarrow$ solve \cref{eq:nonconvexCovTradeoff} with $d_i$ and $\neval=i$ ~\comm{Compute miscoverage level}

  \If{$\varepsilon_i \leq \varepsilon$}
  \comm{Current solution meets target coverage}
    \State $i_\mathrm{val} \leftarrow i$ 
  \comm{Record as the best valid solution}
  \ElsIf{$\xi_m(\qe_i^*) = 0$ for all $m = 1, \ldots, \ncal$}\comm{Coverage insufficient but all constraints satisfied}
    \State \Return None \comm{Return since no valid solution was found}
  \Else \comm{Target coverage not met}
    \State $i_\mathrm{inv} \leftarrow i$ \comm{Record as the best invalid solution}
  \EndIf

  \If{$i_\mathrm{val} \neq 0$} \comm{Valid solution was found previously}
    \State $\varepsilon_{i_\mathrm{val}}' \leftarrow$ \cref{eq:nonconvexCovTradeoff} using $d_{i_\mathrm{val}}$ and $\neval=i+1$ \comm{Check coverage for next iteration}
    \If{$\varepsilon_{i_\mathrm{val}}' > \varepsilon$ \textbf{or} $i \geq i_\mathrm{soft}$}\comm{Will not meet target coverage or soft iteration limit reached}
      \State \Return $q_{i_\mathrm{val}}^*,\;\phi_{i_\mathrm{val}},\;d_{i_\mathrm{val}},\;i_\mathrm{val}$ \comm{Return best valid solution}
    \EndIf
  \EndIf

  \State \vspace{-0.5cm}
  \begin{align*}
      \phi_{i+1} \leftarrow
    \begin{dcases}
      \phi_{i_\mathrm{inv}} \cdot f_+^{\,\max(1,\, i - i_\mathrm{soft})}
        & \text{if } i_\mathrm{val} = 0 ~\text{\comm{Increase best invalid penalty weight}} \\
      \phi_{i_\mathrm{val}} \cdot f_-
        & \text{if } i_\mathrm{inv} = 0  ~\text{\comm{Decrease best valid penalty weight}}\\
      \tfrac{1}{2}(\phi_{i_\mathrm{val}} + \phi_{i_\mathrm{inv}})
        & \text{otherwise} ~\text{\comm{Set to midpoint between best penalty weights}}
    \end{dcases}
  \end{align*}
  \State $i \leftarrow i + 1$ \comm{Next iteration}
\EndWhile
\State \Return None \comm{No valid solution within $i_\mathrm{max}$ iterations}

\end{algorithmic}
\end{algorithm}

\subsection{Practical Design Guidelines}\label{subsec:scoreCostDesign}
The choice of score and cost functions governs the computational complexity of calibration, the tractability of test-time prediction, and the conservatism of the resulting prediction sets.
We discuss the main design strategies here; concrete instantiations for rectangular, ellipsoidal, zonotopic, sublevel, and multi-modal prediction sets are provided in Appendix~\ref{sec:Examples}.

\textbf{Score function design.}
The score function can be designed by specifying a parametrized family of prediction sets $\Y(x,\qe)$ for which set membership can be efficiently evaluated, and define the score function implicitly via $s(x,y,\qe) = \mathbb{I}\{y \notin \Y(x,\qe)\}$, where $\mathbb{I}(\cdot)$ is the indicator function; RemMCP then requires the sets $\{\qe \mid y\in\Y(x,\qe)\}$ to be convex.
This yields a closed-form description of the set boundary.
Alternatively, we can define the prediction set as the sublevel set $\Y(x,\qe) = \{y \mid g(x,y,\qe_1) \leq \qe_2\}$ of an arbitrary function $g$, with score function $s(x, y, \qe) = g(x,y,\qe_1)-\qe_2$, simultaneously calibrating the sublevel threshold $\qe_2$ and additional shape parameters $\qe_1$.
This yields highly flexible prediction set shapes but does not produce a closed-form set boundary.
Unions of prediction sets can be obtained by combining score functions via the $\min$-operator analogously to~\citep{tumu2024MultiModalCP}, with the advantage that all component parameters are optimized and calibrated jointly without requiring an additional data split.

\textbf{Cost function design.} 
The cost function typically penalizes prediction set volume, though other objectives are equally admissible. When a closed-form proxy is unavailable --- as is the case for volume over general sublevel set functions --- Monte Carlo estimates offer a practical alternative. Concretely, we can optimize the empirical quality of the prediction set averaged over a representative set of inputs $\mathcal{X}$, i.e., 
$J(\qe) = \sum_{x\in\mathcal{X}} \ell \bigl(\Y(x,\qe)\bigr)$,
where $\ell(\cdot)$ denotes an appropriate performance measure such as the volume, diameter, or a surrogate thereof.

\textbf{Connection to existing frameworks.}
SCP~\citep{vovk2013condVal}, interval predictor models~\citep{campi2009intervalPred}, zono-conformal prediction~\citep{luetzow2025zonoconformal}, reachset-conformant identification~\citep{Liu2023conf,luetzow2026reachset}, and data-driven reachability analysis~\citep{dietrich2024nonconvex} can all be seen as special cases of MCP (see Appendix~\ref{sec:Examples}).
Their score and cost functions can thus inspire novel MCP instantiations.

\begin{figure}[t]
  \centering
  \sharedlegendSetsVehNew
  \vspace{-0.1cm}
  
\begin{subfigure}[t]{0.35\textwidth}
  \centering
    \includegraphics[page=1, trim=0cm 0.19cm 0cm 0cm, clip]{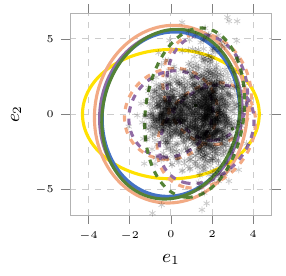}    
  \caption{$T=2$ (residuals scaled by $10^3$).}
  \label{fig:fig_vehicleT2_sets5}
\end{subfigure}%
\hfill
\begin{subfigure}[t]{0.32\textwidth}
  \centering
    \includegraphics[page=2, trim=0cm 0.19cm 0cm 0cm, clip]{figures/fig_environments/fig_vehicle_sets_hc.pdf}
  \caption{$T=30$.}
  \label{fig:fig_vehicle30_sets5}
\end{subfigure}%
\hfill
\begin{subfigure}[t]{0.325\textwidth}
  \centering
    \includegraphics[page=3, trim=0cm 0.19cm 0cm 0cm, clip]{figures/fig_environments/fig_vehicle_sets_hc.pdf}
  \caption{$T=50$.}
  \label{fig:fig_vehicle50_sets5}
\end{subfigure}
\caption{{Vehicle} prediction sets and calibration residuals ($*$) at $\varepsilon=0.05$ and $\beta=10^{-6}$ for run 1.}
  \label{fig:fig_vehicle_sets5}
  \vspace{0.3cm}

  \sharedlegendVehNew
  
  \input{figures/rawTikz_hc/fig_mainHc_Vehicle_boxplot}
\caption{Empirical coverage (violins with quartiles and median) and set volume for {vehicle} data at $\varepsilon=0.05$ and $\beta=10^{-6}$ over 10{,}000 test points and 500 runs. The red dashed line marks the target coverage $1-\varepsilon$, diamonds mark means, and triangles mark mean volumes above the plot range.}
  \label{fig:fig_vehicle_box}
\vspace{-0.5cm}
\end{figure}

\section{Numerical Experiments} \label{sec:experiments}
We empirically evaluate MCP against standard SCP baselines using the vehicle prediction benchmark from~\cite[Sec.~4.2]{tumu2024MultiModalCP}. The point predictor $f(x)$ maps the vehicle state over the past 5 time steps, $x$, to the vehicle position $T$ time steps into the future, $y$ (time-step size $0.1\,\mathrm{s}$). For each run we randomly generate $\ncal = 2{,}000$ calibration points for a vehicle at an intersection. 
We consider the following methods for generating ellipsoidal and multi-modal prediction sets:
\begin{itemize}[leftmargin=*, nosep]
    \item[]\hspace{-0.3cm}\emph{SCP-1:} Standard SCP with score $\tilde{s}(x,y)=\|y-f(x)\|_2$. 
    This score yields spherical prediction sets and serves as the geometry-agnostic baseline.
    \item[]\hspace{-0.3cm}\emph{SCP-2:} SCP in which the $\ncal$ calibration points are split: a held-out subset of $\ncal'=\frac{1}{4}\ncal$ points is reserved for threshold calibration, while the remainder is used to estimate
    \begin{itemize}[leftmargin=1.5em, nosep]
        \item[\emph{A})] the residual covariance $\hat{\Sigma}$ of an ellipsoidal prediction set, via the score $\tilde{s}(x,y)=(y-f(x))^\top \hat{\Sigma}^{-1}(y-f(x))$~\citep{johnstone2021conformal,messoudi2022ellipsoidal}, or
        \item[\emph{B})] ellipsoid centers $\hat{c}_i$ via $k$-means and per-cluster sample covariances $\hat{\Sigma}_i$ of a union-of-ellipsoids prediction region, via the score $\tilde{s}(x,y)=\min_{i\in\{1,2,3\}}(y-f(x)-\hat{c}_i)^\top \hat{\Sigma}_i^{-1}(y-f(x)-\hat{c}_i)$.
    \end{itemize}
    \item[]\hspace{-0.3cm}\emph{SCP-3:} As SCP-2, but reserving $\ncal'=\frac{1}{2}\ncal$ points for threshold calibration.
    \item[]\hspace{-0.3cm}\emph{RemMCP:} The MCP approach via constraint removal (\cref{subsec:RemMCP}), which optimizes the shape of an ellipsoidal prediction set via the score $s(x,y,\qe)=\|L^\top(y-f(x))\|_2-1$ over the full calibration set while minimizing ellipsoid volume, where the optimization variables are the entries of the lower-triangular matrix $L$ with $\Sigma^{-1}=LL^\top$ (see~Appendix~\ref{sec:Examples}). 
    \item[]\hspace{-0.3cm}\emph{H-RelMCP:} The heuristic version of the MCP approach via constraint relaxation (\cref{subsec:RelMCP}), using
    \begin{itemize}[leftmargin=1.5em, nosep]
        \item[\emph{A})] the same convex single-ellipsoid score and cost as RemMCP for a direct comparison, or 
        \item[\emph{B})] the total ellipsoid volume as cost and the non-convex union-of-ellipsoids score $s(x,y,\qe)=\min_{i\in\{1,2,3\}}\|L_i^\top(y-f(x)-c_i)\|_2-1$, where the centers $c_i$ and the lower-triangular matrices $L_i$ with $\Sigma_i^{-1}=L_iL_i^\top$ are jointly optimized during calibration.
    \end{itemize}
\end{itemize}
All methods target the high-confidence guarantee in \cref{eq:confGuarantee}, which is the most desirable guarantee in safety-critical settings, with $\varepsilon = 0.05$ and $\beta = 10^{-6}$. The SCP baselines choose $\rho$ via \cref{eq:noutMCPconf} with $\nq=1$ and $\ncal$ being their number of threshold calibration points, RemMCP via \cref{eq:noutMCPconf}, and H-RelMCP computes its solution via \cref{alg:RelMCP}. Experimental details are provided in Appendix~\ref{sec:experimentalDetails}.

\Cref{fig:fig_vehicle_sets5} shows prediction sets and calibration residuals for one run. At $T=2$, the residuals are compact and nearly unimodal; at $T=30$ and $T=50$, they cluster around three distinct modes. 

\Cref{fig:fig_vehicle_box} reports empirical coverage and prediction set volume across 500 independent calibration runs. Calibration takes under $10\,\mathrm{ms}$ for SCP, about $0.08\,\mathrm{s}$ for RemMCP, $0.5\,\mathrm{s}$ for H-RelMCP-A, and $26$--$59\,\mathrm{s}$ (median) for H-RelMCP-B. 
Among the single-ellipsoid methods, RemMCP yields the smallest prediction sets for all $T$ (e.g., a mean volume of $5.64$ at $T=30$ vs.\ $5.75$ for H-RelMCP-A, $6.82$ for SCP-3A, and $7.55$ for SCP-2A), whereas the SCP-1 balls are the largest at $T=30$ and $T=50$, as they cannot adapt to the elongated spread of the residuals (\cref{fig:fig_vehicle30_sets5,fig:fig_vehicle50_sets5}). For these $T$, the unions of ellipsoids enclose the three residual modes much more tightly and thus yield substantially smaller sets; H-RelMCP-B yields the smallest sets ($2.76$ vs.\ $2.79$ for SCP-3B and $3.33$ for SCP-2B at $T=30$, and $26.8$ vs.\ $28.0$ and $31.5$ at $T=50$). At $T=2$, the volume of H-RelMCP-B remains stable across runs (mean $42.7$), whereas a few degenerate runs inflate the mean volumes of SCP-2B and SCP-3B to $1{,}430$ and $628$.
All methods meet the target coverage in every run (lowest empirical coverage: $0.954$) and the target confidence (misconfidence at most $9.9\cdot10^{-7}$), but the methods differ in how far above it and how much the coverage varies across runs. SCP-1 and RemMCP are closest to the target (mean coverage $0.971$--$0.974$), followed by H-RelMCP ($0.977$--$0.979$) and SCP-3 ($0.979$), whereas SCP-2 is the most conservative ($0.988$). The violins of the split methods SCP-2 and SCP-3 are also wider, with longer tails (standard deviation $0.0045$--$0.0053$ versus $0.0036$--$0.0042$ for SCP-1, RemMCP, and H-RelMCP).

\begin{wrapfigure}[15]{r}{0.4\textwidth}  %
  \centering
  \vspace{-0.5cm}
  \begin{tikzpicture}
\begin{axis}[
    width=6.5cm,
    height=4.0cm,
    xlabel={Coverage Probability $\eta(q)$},
    ylabel={},
    xmin=0.9, xmax=1.0,
    ymin=0,
    ytick=\empty,
    xtick={0.9,0.95,1.0},
    xticklabel style={font=\small},
    xlabel style={font=\small},
    ylabel style={font=\small},
    axis lines=left,
    axis line style={-},
    tick style={color=black, thin},
    clip=false,
    legend style={
        at={(0.,0.97)},
        anchor=north west,
        font=\footnotesize,
        draw=none,
        fill=none,
        row sep=1pt,
    },
    legend cell align=left,
]

\addplot[
    color=scp1color,
    line width=2pt,restrict x to domain=0.9:inf
] table {figures/fig_helpers/beta_coords_ncal2000_eps5_nq1_rho56};
\addlegendentry{SCP-1} %

\addplot[
    color=scp3color,
    line width=1.4pt,restrict x to domain=0.9:inf
] table {figures/fig_helpers/beta_coords_ncal500_eps5_nq1_rho5};
\addlegendentry{{SCP-2}} %

\addplot[
    color=scp4color,
    line width=1.4pt,restrict x to domain=0.9:inf
] table {figures/fig_helpers/beta_coords_ncal1000_eps5_nq1_rho20};
\addlegendentry{{SCP-3}} %

\addplot[
    color=cmcp1color,
    line width=1.2pt,restrict x to domain=0.9:inf
] table {figures/fig_helpers/beta_coords_ncal2000_eps5_nq3_rho18};
\addlegendentry{RemMCP} %

\addplot[
    color=black!40!red,
    dashed,
    line width=1.0pt,
] coordinates {(0.95, 0) (0.95, 118)};
\node[
    anchor=south west,
    font=\footnotesize,
    text=black!30!red,
] at (axis cs: 0.95, 101) {~$1-\varepsilon$};
\end{axis}
\end{tikzpicture}
  \vspace{-0.8cm}
  \caption{Coverage distributions for $\varepsilon=0.05$ and $\beta=10^{-6}$ from \cref{eq:betaMCP} with $\rho$ from \cref{eq:noutMCPconf}. SCP-1 and RemMCP coincide; for RemMCP, the true coverage is stochastically larger in general.}
  \label{fig:fig_betaDistr}
\vspace{-0.3cm}
\end{wrapfigure}
This behavior is accurately predicted by the theoretical coverage distributions derived via~\cref{eq:betaMCP} and visualized in~\cref{fig:fig_betaDistr}.
With only $\ncal'=\frac{1}{4}\ncal$ points for threshold estimation, the coverage distribution of \mbox{SCP-2} has substantially heavier tails than those of SCP-1 or RemMCP, so that SCP-2 needs a much higher mean coverage to reach the same confidence. 
Allocating half the data to calibration (SCP-3) narrows the distribution but correspondingly reduces the data available for shape estimation. Achieving low variance without sacrificing data efficiency requires using all calibration data for shape optimization and threshold calibration simultaneously --- precisely what MCP is designed to do.

Additional results over varying $\varepsilon$, $\beta$, and $\ny$ on further datasets are provided in Appendix~\ref{sec:addResults}: MCP's advantage grows as $\beta$ decreases, whereas both MCP variants become conservative for calibration problems far from fully supported. Under the expected guarantee (Appendix~\ref{sec:expectedResults}), SCP and RemMCP meet the target on average, RemMCP again yields the smallest single-ellipsoid sets at $T=30$ and $T=50$, and a data split increases the coverage standard deviation.

\section{Conclusions}\label{sec:conclusions}
We have introduced multi-variable conformal prediction (MCP), which extends conformal calibration to vector-valued score functions and multiple optimization variables, eliminating the need to split the calibration data between shape optimization and threshold calibration. By grounding MCP in scenario theory, we derived two tractable variants — RemMCP, based on constrained optimization with constraint removal, and RelMCP, based on constraint relaxation — each equipped with finite-sample coverage guarantees under mild assumptions. RemMCP generalizes split conformal prediction and is provably less conservative than its dimension-wise Bonferroni-corrected version. 
Numerical experiments demonstrate that RemMCP and the heuristic, adaptive version H-RelMCP consistently meet the target coverage with lower variance than baselines with data split while producing prediction sets smaller than or comparable to split conformal baselines as long as the calibration problem is close to fully supported. 
Limitations (Appendix~\ref{sec:lim}) include the higher computational cost, the need for a sufficiently large calibration set, and the missing guarantee for H-RelMCP. Future work should address these limitations together with richer score and cost function designs and settings with distribution shift — directions that we expect will further broaden the practical impact of the framework.

\begin{ack}
This work was funded by the Deutsche Forschungsgemeinschaft (DFG, German Research Foundation) – SFB 1608 – 501798263.

\end{ack}

\bibliographystyle{plainnat}
\bibliography{literature}

\begin{thebibliography}{48}
\providecommand{\natexlab}[1]{#1}
\providecommand{\url}[1]{\texttt{#1}}
\expandafter\ifx\csname urlstyle\endcsname\relax
  \providecommand{\doi}[1]{doi: #1}\else
  \providecommand{\doi}{doi: \begingroup \urlstyle{rm}\Url}\fi

\bibitem[Angelopoulos and Bates(2022)]{angelopoulos2022gentleintroductionCP}
Anastasios~N. Angelopoulos and Stephen Bates.
\newblock A gentle introduction to conformal prediction and distribution-free uncertainty quantification.
\newblock \emph{Preprint}, 2022.
\newblock https://arxiv.org/abs/2107.07511.

\bibitem[Angelopoulos et~al.(2026)Angelopoulos, Barber, and Bates]{angelopoulos2026theoreticalfoundationsCP}
Anastasios~N. Angelopoulos, Rina~Foygel Barber, and Stephen Bates.
\newblock \emph{Theoretical Foundations of Conformal Prediction}.
\newblock Cambridge University Press, 2026.
\newblock Preprint, https://arxiv.org/abs/2411.11824.

\bibitem[Bai et~al.(2022)Bai, Mei, Wang, Zhou, and Xiong]{bai2022efficientDifferentiableCP}
Yu~Bai, Song Mei, Huan Wang, Yingbo Zhou, and Caiming Xiong.
\newblock Efficient and differentiable conformal prediction with general function classes.
\newblock In \emph{International Conference on Learning Representations}, 2022.

\bibitem[Bowman and Heath(2023)]{bowman2023mvee}
Nathaniel Bowman and Michael~T. Heath.
\newblock Computing minimum-volume enclosing ellipsoids.
\newblock \emph{Mathematical Programming Computation}, 15\penalty0 (4):\penalty0 621--650, 2023.

\bibitem[Braun et~al.(2026)Braun, Aolaritei, Jordan, and Bach]{braun2025minimumvolumeconformalsets}
Sacha Braun, Liviu Aolaritei, Michael~I. Jordan, and Francis Bach.
\newblock Minimum volume conformal sets for multivariate regression.
\newblock \emph{Journal of the American Statistical Association}, pages 1--27, 2026.

\bibitem[Calafiore and Campi(2005)]{calafiore2005uncertain}
Giuseppe~C. Calafiore and Marco~C. Campi.
\newblock Uncertain convex programs: randomized solutions and confidence levels.
\newblock \emph{Mathematical Programming}, 102\penalty0 (1):\penalty0 25--46, 2005.

\bibitem[Calafiore and Campi(2006)]{calafiore2006scenario}
Giuseppe~C. Calafiore and Marco~C. Campi.
\newblock The scenario approach to robust control design.
\newblock \emph{IEEE Transactions on Automatic Control}, 51\penalty0 (5):\penalty0 742--753, 2006.

\bibitem[Campi and Garatti(2008)]{campi2008exact}
Marco~C. Campi and Simone Garatti.
\newblock The exact feasibility of randomized solutions of uncertain convex programs.
\newblock \emph{SIAM Journal on Optimization}, 19\penalty0 (3):\penalty0 1211--1230, 2008.

\bibitem[Campi and Garatti(2011)]{campi2011samplingDiscard}
Marco~C. Campi and Simone Garatti.
\newblock A sampling-and-discarding approach to chance-constrained optimization: {F}easibility and optimality.
\newblock \emph{Journal of Optimization Theory and Applications}, 148:\penalty0 257--280, 2011.

\bibitem[Campi et~al.(2009)Campi, Calafiore, and Garatti]{campi2009intervalPred}
Marco~C. Campi, Giuseppe~C. Calafiore, and Simone Garatti.
\newblock Interval predictor models: {I}dentification and reliability.
\newblock \emph{Automatica}, 45\penalty0 (2):\penalty0 382--392, 2009.

\bibitem[Campi et~al.(2015)Campi, Garatti, and Ramponi]{campi2015nonconvex}
Marco~C Campi, Simone Garatti, and Federico~A Ramponi.
\newblock Non-convex scenario optimization with application to system identification.
\newblock In \emph{IEEE Conference on Decision and Control}, pages 4023--4028, 2015.

\bibitem[Campi et~al.(2021)Campi, Car\`e, and Garatti]{campi2021scenario}
Marco~C. Campi, Algo Car\`e, and Simone Garatti.
\newblock The scenario approach: a tool at the service of data-driven decision making.
\newblock \emph{Annual Reviews in Control}, 52:\penalty0 1--17, 2021.

\bibitem[Cini et~al.(2025)Cini, Jenkins, Mandic, Alippi, and Bianchi]{cini2025relational}
Andrea Cini, Alexander Jenkins, Danilo Mandic, Cesare Alippi, and Filippo~Maria Bianchi.
\newblock Relational conformal prediction for correlated time series.
\newblock In \emph{International Conference on Machine Learning}, 2025.

\bibitem[Cleaveland et~al.(2024)Cleaveland, Lee, Pappas, and Lindemann]{cleaveland2024conformalLCP}
Matthew Cleaveland, Insup Lee, George~J. Pappas, and Lars Lindemann.
\newblock Conformal prediction regions for time series using linear complementarity programming.
\newblock In \emph{AAAI Conference on Artificial Intelligence}, 2024.

\bibitem[Coppola et~al.(2024)Coppola, Peruffo, Lindemann, and Mazo]{coppola2024scenarioCP}
Rudi Coppola, Andrea Peruffo, Lars Lindemann, and Manuel Mazo.
\newblock Scenario approach and conformal prediction for verification of unknown systems via data-driven abstractions.
\newblock In \emph{European Control Conference}, pages 558--563, 2024.

\bibitem[de~Angelis et~al.(2021)de~Angelis, Rocchetta, Gray, and Ferson]{deAngelis2021consonantPredictiveBeliefs}
Marco de~Angelis, Roberto Rocchetta, Ander Gray, and Scott Ferson.
\newblock Constructing consonant predictive beliefs from data with scenario theory.
\newblock In \emph{International Symposium on Imprecise Probabilities and Their Applications}, 2021.

\bibitem[Dietrich et~al.(2024)Dietrich, Devonport, and Arcak]{dietrich2024nonconvex}
Elizabeth Dietrich, Alex Devonport, and Murat Arcak.
\newblock Nonconvex scenario optimization for data-driven reachability.
\newblock In \emph{Conference on Learning for Dynamics and Control}, pages 514--527, 2024.

\bibitem[Fang et~al.(2025)Fang, Tan, and Huang]{fang2025contra}
Zhenhan Fang, Aixin Tan, and Jian Huang.
\newblock {CONTRA}: {C}onformal prediction region via normalizing flow transformation.
\newblock In \emph{International Conference on Learning Representations}, 2025.

\bibitem[Feldman et~al.(2023)Feldman, Bates, and Romano]{feldman2023multiQuantile}
Shai Feldman, Stephen Bates, and Yaniv Romano.
\newblock Calibrated multiple-output quantile regression with representation learning.
\newblock \emph{Journal of Machine Learning Research}, 24\penalty0 (24):\penalty0 1--48, 2023.

\bibitem[Garatti and Campi(2022)]{garatti2022risk}
Simone Garatti and Marco~C. Campi.
\newblock Risk and complexity in scenario optimization.
\newblock \emph{Mathematical Programming}, 191\penalty0 (1):\penalty0 243--279, 2022.

\bibitem[Garatti and Campi(2025)]{garatti2024nonconvex}
Simone Garatti and Marco~C. Campi.
\newblock Non-convex scenario optimization.
\newblock \emph{Mathematical Programming}, 209\penalty0 (1):\penalty0 557–608, 2025.

\bibitem[Garatti et~al.(2019)Garatti, Campi, and Car\`{e}]{garatti2019interval}
Simone Garatti, Marco~C. Campi, and Algo Car\`{e}.
\newblock On a class of interval predictor models with universal reliability.
\newblock \emph{Automatica}, 110\penalty0 (C), 2019.
\newblock article no. 108542.

\bibitem[Gray et~al.(2025)Gray, Gopakumar, Rousseau, and Destercke]{gray2025guaranteed}
Ander Gray, Vignesh Gopakumar, Sylvain Rousseau, and Sebastien Destercke.
\newblock Guaranteed prediction sets for functional surrogate models.
\newblock In \emph{Conference on Uncertainty in Artificial Intelligence}, 2025.

\bibitem[Johnstone and Cox(2021)]{johnstone2021conformal}
Chancellor Johnstone and Bruce Cox.
\newblock Conformal uncertainty sets for robust optimization.
\newblock In \emph{Proceedings of the 10th Symposium on Conformal and Probabilistic Prediction and Applications}, volume 152, pages 72--90, 2021.

\bibitem[Klein et~al.(2026)Klein, B{\'e}thune, Ndiaye, and Cuturi]{klein2025CPoptimalTransport}
Michal Klein, Louis B{\'e}thune, Eugene Ndiaye, and Marco Cuturi.
\newblock Multivariate conformal prediction using optimal transport.
\newblock \emph{Transactions on Machine Learning Research}, 2026.

\bibitem[Lei et~al.(2013)Lei, Robins, and Wasserman]{Lei2013distributionfree}
Jing Lei, James Robins, and Larry Wasserman.
\newblock Distribution-free prediction sets.
\newblock \emph{Journal of the American Statistical Association}, 108\penalty0 (501):\penalty0 278--287, 2013.

\bibitem[Lin and Bansal(2024)]{lin2024verificationScenarioCP}
Albert Lin and Somil Bansal.
\newblock Verification of neural reachable tubes via scenario optimization and conformal prediction.
\newblock In \emph{Conference on Learning for Dynamics and Control}, volume 242, pages 719--731, 2024.

\bibitem[Liu et~al.(2023)Liu, Schürmann, and Althoff]{Liu2023conf}
Stefan~B. Liu, Bastian Schürmann, and Matthias Althoff.
\newblock Guarantees for real robotic systems: {U}nifying formal controller synthesis and reachset-conformant identification.
\newblock \emph{IEEE Transactions on Robotics}, 39\penalty0 (5):\penalty0 3776--3790, 2023.

\bibitem[Luo and Zhou(2026)]{luo2026densitysortedpredictionsetefficient}
Rui Luo and Zhixin Zhou.
\newblock Density-sorted prediction set: Efficient conformal prediction for multi-target regression.
\newblock \emph{Pattern Recognition}, 172, 2026.
\newblock article no. 112513.

\bibitem[L\"utzow and Althoff(2024)]{luetzow2024generator}
Laura L\"utzow and Matthias Althoff.
\newblock Scalable reachset-conformant identification of linear systems.
\newblock \emph{IEEE Control Systems Letters}, 8:\penalty0 520--525, 2024.

\bibitem[Lützow and Althoff(2026)]{luetzow2026reachset}
Laura Lützow and Matthias Althoff.
\newblock Reachset-conformant system identification.
\newblock \emph{IEEE Transactions on Automatic Control}, 71\penalty0 (5):\penalty0 3014--3029, 2026.

\bibitem[Lützow et~al.(2025)Lützow, Eichelbeck, Kochenderfer, and Althoff]{luetzow2025zonoconformal}
Laura Lützow, Michael Eichelbeck, Mykel Kochenderfer, and Matthias Althoff.
\newblock Zono-conformal prediction: {Z}onotope-based uncertainty quantification for regression and classification tasks.
\newblock \emph{Journal of Machine Learning Research}, 26\penalty0 (294):\penalty0 1--37, 2025.

\bibitem[Messoudi et~al.(2021)Messoudi, Destercke, and Rousseau]{messoudi2021copula}
Soundouss Messoudi, Sébastien Destercke, and Sylvain Rousseau.
\newblock Copula-based conformal prediction for multi-target regression.
\newblock \emph{Pattern Recognition}, 120, 2021.
\newblock article no. 108101.

\bibitem[Messoudi et~al.(2022)Messoudi, Destercke, and Rousseau]{messoudi2022ellipsoidal}
Soundouss Messoudi, S\'{e}bastien Destercke, and Sylvain Rousseau.
\newblock Ellipsoidal conformal inference for multi-target regression.
\newblock In \emph{Symposium on Conformal and Probabilistic Prediction with Applications}, volume 179, pages 294--306, 2022.

\bibitem[O'Sullivan et~al.(2025)O'Sullivan, Romao, and Margellos]{osullivan2025bridgingconformalpredictionscenario}
Niall O'Sullivan, Licio Romao, and Kostas Margellos.
\newblock Bridging conformal prediction and scenario optimization.
\newblock In \emph{IEEE Conference on Decision and Control}, pages 6114--6121, 2025.

\bibitem[Park and Cho(2025)]{park2025semiparametricCP}
Ji~Won Park and Kyunghyun Cho.
\newblock Semiparametric conformal prediction.
\newblock In \emph{International Conference on Artificial Intelligence and Statistics}, 2025.

\bibitem[Romao et~al.(2023{\natexlab{a}})Romao, Margellos, and Papachristodoulou]{romao2023probabilistic}
Licio Romao, Kostas Margellos, and Antonis Papachristodoulou.
\newblock Probabilistic feasibility guarantees for convex scenario programs with an arbitrary number of discarded constraints.
\newblock \emph{Automatica}, 149, 2023{\natexlab{a}}.
\newblock article no. 110601.

\bibitem[Romao et~al.(2023{\natexlab{b}})Romao, Papachristodoulou, and Margellos]{romao2023exactFeasibility}
Licio Romao, Antonis Papachristodoulou, and Kostas Margellos.
\newblock On the exact feasibility of convex scenario programs with discarded constraints.
\newblock \emph{IEEE Transactions on Automatic Control}, 68\penalty0 (4):\penalty0 1986--2001, 2023{\natexlab{b}}.

\bibitem[Sadeghi et~al.(2019)Sadeghi, {de Angelis}, and Patelli]{sadeghi2019efficientTraining}
Jonathan Sadeghi, Marco {de Angelis}, and Edoardo Patelli.
\newblock Efficient training of interval neural networks for imprecise training data.
\newblock \emph{Neural Networks}, 118:\penalty0 338--351, 2019.

\bibitem[Spyromitros-Xioufis et~al.(2016)Spyromitros-Xioufis, Tsoumakas, Groves, and Vlahavas]{spyromitros2016rf1}
Eleftherios Spyromitros-Xioufis, Grigorios Tsoumakas, William Groves, and Ioannis Vlahavas.
\newblock Multi-target regression via input space expansion: {T}reating targets as inputs.
\newblock \emph{Machine Learning}, 104:\penalty0 55--98, 2016.

\bibitem[Stankevi{\v{c}}i{\={u}}t{\.{e}} et~al.(2021)Stankevi{\v{c}}i{\={u}}t{\.{e}}, Alaa, and van~der Schaar]{stankeviciute2021conformal}
Kamil{\.{e}} Stankevi{\v{c}}i{\={u}}t{\.{e}}, Ahmed Alaa, and Mihaela van~der Schaar.
\newblock Conformal time-series forecasting.
\newblock In \emph{Advances in Neural Information Processing Systems}, 2021.
\newblock article no. 475.

\bibitem[Sun and Yu(2024)]{sun2024copula}
Sophia~Huiwen Sun and Rose Yu.
\newblock Copula conformal prediction for multi-step time series prediction.
\newblock In \emph{International Conference on Learning Representations}, 2024.

\bibitem[Tawachi and Laufer-Goldshtein(2025)]{tawachi2025multidimensionalCP}
Yam Tawachi and Bracha Laufer-Goldshtein.
\newblock Multi-dimensional conformal prediction.
\newblock In \emph{International Conference on Learning Representations}, 2025.

\bibitem[Tumu et~al.(2024)Tumu, Cleaveland, Mangharam, Pappas, and Lindemann]{tumu2024MultiModalCP}
Renukanandan Tumu, Matthew Cleaveland, Rahul Mangharam, George~J. Pappas, and Lars Lindemann.
\newblock Multi-modal conformal prediction regions by optimizing convex shape templates.
\newblock In \emph{Conference on Learning for Dynamics and Control}, 2024.

\bibitem[Vovk(2013)]{vovk2013condVal}
Vladimir Vovk.
\newblock Conditional validity of inductive conformal predictors.
\newblock \emph{Machine Learning}, 92\penalty0 (2–3):\penalty0 349--376, 2013.

\bibitem[Vovk et~al.(1999)Vovk, Gammerman, and Saunders]{vovk1999randomness}
Volodya Vovk, Alexander Gammerman, and Craig Saunders.
\newblock Machine-learning applications of algorithmic randomness.
\newblock In \emph{International Conference on Machine Learning}, pages 444--453, 1999.

\bibitem[Xu et~al.(2024)Xu, Jiang, and Xie]{xu2024cpMulti}
Chen Xu, Hanyang Jiang, and Yao Xie.
\newblock Conformal prediction for multi-dimensional time series by ellipsoidal sets.
\newblock In \emph{International Conference on Machine Learning}, 2024.
\newblock article no. 2268.

\bibitem[Zhang et~al.(2025)Zhang, Li, and Bastani]{zhang2025confStructured}
Botong Zhang, Shuo Li, and Osbert Bastani.
\newblock Conformal structured prediction.
\newblock In \emph{International Conference on Learning Representations}, pages 8563--8583, 2025.

\end{thebibliography}

\newpage
\appendix

\section{Proofs}\label{sec:proofs}
This section provides the proofs of the theorems and the corollaries of the main paper.

\subsection{Proof of \texorpdfstring{\cref{thm:multiVarCP}}{Theorem \getrefnumber{thm:multiVarCP}}}
The programs in \cref{eq:MCPprogram} are scenario programs of the form \cref{eq:SP} with $\delta=(x,y)$, convex sets $\mathcal{Q}_{\delta}=\{\qe\colon s(x,y,\qe) \leq \mathbf{0}\}$, and convex objective $J(\qe)$, where \cref{assumption:convex} holds. If additionally \cref{assumption:unique,assumption:nondenerate} hold and the calibration data points and the test point are exchangeable (as required by \cref{thm:multiVarCP}), \cref{eq:expCovScenario} applies. Since the coverage satisfies $\eta(\qe) = 1 - V(\qe)$ for any $\qe$, where $V(\qe)$ denotes the violation probability of the corresponding scenario program, this yields
\begin{align*}
    \mathbb{E}[\eta(\qe^*)] = 1-\mathbb{E}[V(\qe^*)] \geq 1-\frac{\nq(\rho+1)}{\ncal+1},
\end{align*}
which establishes \cref{eq:expGuarantee} with $\alpha=\nq(\rho+1)/(\ncal+1)$ for any $\rho\in\mathbb{N}_0$. For $\rho$ chosen via \cref{eq:noutMCP}, we have $\nq(\rho+1)/(\ncal+1)\leq\alpha$ and hence $\mathbb{E}[\eta(\qe^*)]\geq 1-\alpha$. If the samples are i.i.d., the high-confidence bound in \cref{eq:confGuarantee} with $\beta=\bar{\beta}(\rho)$ given by \cref{eq:noutMCPconf} follows directly from \cref{eq:varCovScenario}, since $\eta(\qe^*)\geq 1-\varepsilon$ if and only if $V(\qe^*)\leq\varepsilon$. For $\rho$ chosen via \cref{eq:noutMCPconf}, we have $\bar{\beta}(\rho)\leq\beta$, so that \cref{eq:confGuarantee} holds for the target $\beta$. 
Under the additional \cref{assumption:fullSupport,assumption:scenarioViolation}, \cref{eq:varCovScenario} holds with equality \citep[Thm.~5]{romao2023exactFeasibility}, and hence so does the high-confidence bound in \cref{eq:confGuarantee} for RemMCP, which yields \cref{eq:betaMCP}.

\subsection{Proof of \texorpdfstring{\cref{cor:RemMCP2scp}}{Corollary \getrefnumber{cor:RemMCP2scp}}}
We first show that SCP and RemMCP yield identical prediction sets, and then derive the coverage guarantees of SCP from those of RemMCP.

\textbf{Prediction Sets.} For an arbitrary SCP score function $\tilde{s}(x,y)$, the SCP prediction set defined in \cref{eq:YCP} coincides with the MCP prediction set in \cref{eq:YMCP} with score function \cref{eq:MCP2CPscore} if $\qe^*=\tilde{\qe}^*$. Hence, it suffices to show that the parameter $\qe^*$ obtained from the RemMCP procedure with \cref{eq:MCP2CP} equals the SCP threshold $\tilde{\qe}^*$.

As in SCP, the calibration procedure of RemMCP with $\rho$ from \cref{eq:noutMCP} begins by computing the outlier budget for a given target miscoverage level $\alpha$.
For $\nq=1$, the outlier budget in \cref{eq:noutMCP} reduces to $\rho=\lfloor\alpha(\ncal+1)\rfloor-1$, which coincides with the SCP budget in \cref{eq:noutCP}.
The optimal parameter $\qe^*$ in RemMCP is obtained by solving the cascading optimization problem in \cref{eq:MCPprogram}. 
For the score and cost functions in \cref{eq:MCP2CP}, which are convex in $\qe$ and therefore admissible for RemMCP, this reduces to
\begin{align}\label{eq:MCPprogram_SCP}
    \qe_k^* = \argmin_{\qe\in\mathbb{R}^{\nq}}~ \qe \quad 
    \text{s.t.}~~  \tilde{s}(x^{(m)},y^{(m)})\leq\qe,
    ~~ \forall (x,y)\in \Mcal\setminus\Moutk{k}. 
\end{align}
At each stage $k=0,\dots,\rho$, the optimizer $\qe_k^*$ of \cref{eq:MCPprogram_SCP} equals the maximum nonconformity score $\tilde{s}(x^{(m)},y^{(m)})$ over the samples in $\Mcal\setminus\Moutk{k}$.
By \cref{eq:Moutk} with $\nq=1$, the set of discarded samples $\Moutk{k}$ at stage $k$ is constructed by adding the current support scenario to the previously discarded samples in $\Moutk{k-1}$.
The support sample at stage $k$ is the data point whose removal would alter the optimizer $\qe_k^*$, and under a continuous score distribution, the maximum nonconformity score is attained uniquely with probability one; hence, exactly one data point, the one corresponding to the maximum nonconformity score, serves as the support sample at each stage.
Consequently, after $\rho$ stages, the set $\Moutk{\rho}$ contains the $\rho$ samples with the largest nonconformity scores, and $\qe^*=\qe_{\rho}^*$ equals the $(\ncal-\rho)$-smallest nonconformity score. This coincides exactly with the SCP threshold $\tilde{\qe}^*$. Therefore, the prediction sets of SCP and RemMCP are identical.

\textbf{Coverage guarantees.}
Next, we derive the coverage guarantees of SCP in \cref{eq:YCP_cov,eq:YCP_beta} from those of RemMCP.
Since the score is strictly monotone in $\qe$, the optimizer of \cref{eq:MCPprogram} over any finite sample set exists and is unique (\cref{assumption:unique}). Moreover, when the score distribution is continuous, the program is nondegenerate with probability one (\cref{assumption:nondenerate}).
Assuming exchangeability of the calibration data points and the test point, the conditions of \cref{thm:multiVarCP} are satisfied, yielding the expected coverage guarantee in \cref{eq:expGuarantee} for RemMCP, which matches \cref{eq:YCP_cov}.

When $\nq=1$, the distributional result for the coverage of RemMCP in \cref{eq:betaMCP} reduces to $\eta(\qe^*)\sim\mathrm{Beta}(\ncal-\rho,\rho+1)$, which coincides with \cref{eq:YCP_beta}. Both results require i.i.d. calibration data. The additional assumptions for \cref{eq:betaMCP} (\cref{assumption:fullSupport,assumption:scenarioViolation}) hold almost surely in this setting, since the maximum is attained uniquely and each removed score strictly exceeds the remaining threshold with probability one under a continuous score distribution. The continuity condition is identical to that required for \cref{eq:YCP_beta}.

\subsection{Proof of \texorpdfstring{\cref{cor:dSCP}}{Corollary \getrefnumber{cor:dSCP}}} 

With the score function in \cref{eq:MCPdSCPscore}, the MCP prediction set in \cref{eq:YMCP} is an axis-aligned hyperrectangle centered at $f(x)$ with half-width vector $\qe^*$.
Similarly, dimension-wise SCP with the score function in \cref{eq:CPdimScore} produces hyperrectangular prediction sets centered at $f(x)$, where the half-width $\tqe_j^*$ in each dimension $j$ is calibrated independently using standard SCP.
The MCP prediction set is contained in (or equal to) the dimension-wise SCP set if and only if $\qe^*_{(j)} \leq \tqe_j^*$ for all $j=1,\dots,\ny$. We show that this relation holds while both methods target the same expected coverage level $1-\alpha$.

To guarantee joint coverage of $1-\alpha$, dimension-wise SCP calibrates each marginal half-width $\tqe_j^*$ at level $\alpha/\ny$ via Bonferroni correction. By \cref{eq:noutCP}, this yields an outlier budget of $\rho=\lfloor (\alpha/\ny)(\ncal+1)\rfloor-1$ per dimension, which coincides with the RemMCP budget $\rho$ in \cref{eq:noutMCP} for target miscoverage $\alpha$ and $\nq=\ny$. 
Dimension-wise SCP discards, for each dimension $j$, the $\rho$ samples with the largest deviations $|y_{(j)}-f_{(j)}(x)|$. Let $\tilde{\mathcal{R}}_{\rho,j}$ denote the set of samples discarded in dimension $j$, and define $\tilde{\mathcal{R}}_{\rho}=\bigcup_{j=1}^{\ny} \tilde{\mathcal{R}}_{\rho,j}$.
Since dimensions are treated independently, the same sample may be discarded in multiple dimensions, implying $|\tilde{\mathcal{R}}_{\rho}| \leq \rho \ny$.
In contrast, the RemMCP discard set $\Moutk{\rho}$ contains exactly $\rho \ny$ samples. We now show that $\Moutk{\rho} \supseteq \tilde{\mathcal{R}}_{\rho}$.

With the score and cost functions in \cref{eq:MCPdSCP}, the RemMCP optimization problem becomes
\begin{align*}
    \qe_k^* = \argmin_{\qe\in\mathbb{R}^{\nq}}~ \mathbf{1}^\top \qe \quad 
    \text{s.t.}~~ 
        |y-f(x)|
     \leq \qe
    ~~ \forall (x,y)\in \Mcal\setminus\Moutk{k}. 
\end{align*}
Thus, for each dimension $j$, the component $\qe_{k(j)}^*$ equals the maximum deviation $|y_{(j)}-f_{(j)}(x)|$ over the samples in $\Mcal\setminus\Moutk{k}$.
Under a continuous score distribution, this maximum is attained uniquely in each dimension with probability one. The corresponding samples form the support set $\supp(\qe_k^*)$ and are added to the discard set $\Moutk{k+1}$ at the next stage (cf. \cref{eq:Moutk}). 
After $\rho$ stages, $\Moutk{\rho}$ therefore contains, for each dimension $j$, the $\rho$ samples with the largest values of $|y_{(j)}-f_{(j)}(x)|$. Hence, $\Moutk{\rho} \supseteq \tilde{\mathcal{R}}_{\rho}$.

To illustrate this, consider a sample $\tilde{m}$ that attains the largest deviation in dimension $j$ and the second-largest deviation in dimension $j' \neq j$. If $\rho \geq 2$, dimension-wise SCP discards this sample in both dimensions, resulting in $|\tilde{\mathcal{R}}_{\rho}| < \rho \ny$.
In contrast, RemMCP removes this sample already at stage $k=1$ (as the support constraint for dimension $j$), and can subsequently discard an additional sample corresponding to the next-largest deviation in dimension $j'$. This would lead to a strictly larger set of discarded samples.

Since $\Moutk{\rho} \supseteq \tilde{\mathcal{R}}_{\rho}$, the overall solution $\tqe^*$ from dimension-wise SCP is feasible for \cref{eq:MCPprogram}. RemMCP can therefore exploit the remaining outlier budget to further reduce the half-widths, yielding $\qe^*_{(j)} \leq \tqe_j^*$ for all $j=1,\dots,\ny$. Consequently, $\Y(x,\qe^*) \subseteq \YCP(x,\tqe_1)\times \cdots \times \YCP(x,\tqe_{\ny})$ for every $x$, i.e., the prediction sets produced by RemMCP are contained in (or equal to) those from dimension-wise SCP, while both achieve the target expected coverage level $1-\alpha$.

\subsection{Proof of \texorpdfstring{\cref{thm:multiVarCP_nonconvex}}{Theorem \getrefnumber{thm:multiVarCP_nonconvex}}}
By introducing additional optimization variables $\bar{\xi}_m \geq \mathbf{0}$, which compute the slack $\xi_m(\qe)$, problem \cref{eq:RelMCPprogram} is equivalent to
\begin{subequations}\label{eq:RelMCPprogram_constr_full}
    \begin{align}
    \qe_i^* = \argmin_{\qe\in\mathbb{R}^{\nq},\bar{\xi}\in\mathbb{R}_{\geq 0}^{\ncal}}~ &J(\qe) +\phi_i \sum_{m=1}^{\ncal} \bar{\xi}_m \\
    \text{s.t.}~~  s(x^{(m)},y^{(m)},\qe)&\leq \bar{\xi}_m \mathbf{1},
    ~~ \forall m=1,\dots,\ncal, \label{eq:RelMCPprogram_constr}
    \end{align}
\end{subequations}
with $\bar{\xi} = [\bar{\xi}_1~\dots~\bar{\xi}_{\ncal}]^\top$.
This is a non-convex scenario program with constraint relaxation, as introduced in \citet[Sec.~4.1]{garatti2024nonconvex}.  
Since the penalty parameters are fixed before calibration, each of the $\neval$ programs is such a scenario program, and applying \citet[Thm.~10]{garatti2024nonconvex} to the $i$-th program yields
\begin{align*}
    \mathbb{P}\left(\eta(\qe_i^*) \geq 1-\varepsilon_i\right) \geq 1-\frac{\beta}{\neval},  
\end{align*}
where $\varepsilon_i$ is the unique solution of \cref{eq:nonconvexCovTradeoff}; since $\varepsilon_i$ increases with $d_i$, this bound continues to hold if $d_i$ is replaced by an upper bound. 
We evaluate $\neval$ distinct penalty parameters, and since any evaluation may be incorrect with probability $\beta/\neval$, all evaluations and, thus, also the selection are simultaneously guaranteed with confidence $1-\beta$; see also \citet[Footnote~29]{garatti2024nonconvex}. 
Thus, the solution must satisfy $\mathbb{P}\left(\eta(\qe_{i}^*) \geq 1-\varepsilon_{i}\right)    \geq 1-\beta$. For $\varepsilon_{i} \leq \varepsilon$, we obtain
\begin{align*}
    \mathbb{P}\left(\eta(\qe_{i}^*) \geq 1-\varepsilon\right)
    \geq \mathbb{P}\left(\eta(\qe_{i}^*) \geq 1-\varepsilon_{i}\right)
    \geq 
    1- \beta.
\end{align*}
This establishes \cref{eq:confGuarantee}.

\section{Heuristic RelMCP: Adaptive Calibration Algorithm}\label{sec:RelMCPalg}
\cref{alg:RelMCP} describes Heuristic RelMCP (H-RelMCP), an iterative procedure for finding a suitable penalty parameter $\phi_i$ for the unconstrained optimization problem in~\cref{eq:RelMCPprogram} of RelMCP. 
The penalty parameter governs a fundamental trade-off: a large $\phi_i$ penalizes constraint violations heavily, producing solutions that satisfy more calibration constraints (lower $\xi_m$), which in turn decreases the solution complexity $d_i$ and tightens the achievable coverage guarantee; a small $\phi_i$ permits more violations, yielding a higher complexity $d_i$ and looser coverage guarantees, but also less conservative solutions. The goal of the algorithm is to find the value of $\phi_i$ that yields the best (least conservative) prediction sets for the target miscoverage level $\varepsilon$ and the confidence level $1-\beta$. 
We provide a detailed step-by-step explanation of the proposed algorithm below.

\textbf{Initialization.}
The algorithm maintains two iteration counters: $i_\mathrm{val}$, indexing the best {valid} solution found so far --- the one with the smallest penalty $\phi_i$ such that the miscoverage $\varepsilon_i \leq \varepsilon$ --- and $i_\mathrm{inv}$, indexing the best {invalid} solution --- the one with the largest penalty $\phi_i$ for which $\varepsilon_i > \varepsilon$ still holds. Both counters are initialized to zero in Line 1, indicating that no solution has yet been evaluated.

\textbf{Main loop.}
At each iteration $i$, the algorithm solves the penalized program~\cref{eq:RelMCPprogram} with the current penalty $\phi_i$ in Line 3, obtaining a calibrated parameter $q_i^*$, and approximates its solution complexity $d_i$ heuristically in Line 4 by the number $v_i$ of violated calibration points plus $\min(\nq, \ncal - v_i)$. The miscoverage $\varepsilon_i$ is then computed in Line 5 from~\cref{eq:nonconvexCovTradeoff} using $d_i$ and $n_\mathrm{eval} = i$, where the dependence on $i$ reflects a union bound over all iterations evaluated so far (cf.~\cref{thm:multiVarCP_nonconvex}).

Three outcomes are possible:
\begin{enumerate}
    \item \emph{Valid solution} ($\varepsilon_i \leq \varepsilon$): The current solution meets the target coverage guarantee and is recorded as the new best valid solution in Line 7. By the penalty update rule in Line 16, all subsequent penalties satisfy $\phi_j < \phi_i$ for $j > i$, so that, for exact minimizers of~\cref{eq:RelMCPprogram}, future iterations can only yield a less conservative (i.e., better) valid solution or become invalid.
    \item \emph{All constraints satisfied, coverage insufficient} ($\varepsilon_i > \varepsilon$ and $\xi_m = 0$ for all $m$): All calibration constraints are already satisfied with zero slack, so no further increase in $\phi$ can reduce the solution complexity $d_i$ or improve the coverage guarantee. The algorithm returns \texttt{None} in Line 9, indicating that no valid solution was found for the given calibration set size and target level $\varepsilon$.
    \item \emph{Invalid solution} ($\varepsilon_i > \varepsilon$ with some $\xi_m > 0$): The coverage guarantee is not yet met, but active slack variables indicate that increasing $\phi$ may reduce violations and lower $d_i$. The iteration is recorded as the new best invalid solution in Line 11. By the update rule in Line 16, all subsequent penalties satisfy $\phi_j > \phi_i$ for $j > i$, so that, for exact minimizers of~\cref{eq:RelMCPprogram}, future iterations can only yield a more conservative (i.e., better) invalid solution or become valid.
\end{enumerate}
Before continuing, the algorithm checks whether the current best valid solution would remain valid in the next iteration by computing the coverage $\varepsilon_{i_\mathrm{val}}'$ via~\cref{eq:nonconvexCovTradeoff} with $n_\mathrm{eval} = i + 1$ in Line 13. If $\varepsilon_{i_\mathrm{val}}' > \varepsilon$, returning immediately is necessary, since the confidence budget is too tight to validate this solution in any future iteration. If the soft iteration limit $i_\mathrm{soft}$ is reached, the algorithm likewise returns the best valid solution. Otherwise, it continues searching for a smaller $\phi$, retaining $i_\mathrm{val}$ as a fallback if the next iteration is invalid. If no valid solution has been found after $i_\mathrm{max}$ iterations, the loop in Line 2 terminates and the algorithm returns \texttt{None} in Line 18, which bounds the computation time.

After each iteration, $\phi$ is updated in Line 16 according to a bisection-like scheme, guided by the history of valid and invalid solutions:
\begin{itemize}
    \item \emph{No valid solution yet} ($i_\mathrm{val} = 0$): The penalty is increased by multiplying the last invalid penalty $\phi_{i_\mathrm{inv}}$ by $f_+^{\max(1,\, i - i_\mathrm{soft})}>1$, with the exponent growing beyond one once $i$ exceeds $i_\mathrm{soft}$ to accelerate the search before the maximum number of iterations $i_\mathrm{max}$ is reached.
    \item \emph{No invalid solution yet} ($i_\mathrm{inv} = 0$): The penalty is decreased by multiplying the last valid penalty $\phi_{i_\mathrm{val}}$ by $f_- \in (0,1)$, probing whether a smaller $\phi$ can still certify the target guarantee.
    \item \emph{Both valid and invalid solutions exist}: The penalty is set to the midpoint $\tfrac{1}{2}(\phi_{i_\mathrm{val}} + \phi_{i_\mathrm{inv}})$, performing binary search between the best known valid and invalid penalties.
\end{itemize}

\textbf{Termination.}
The algorithm returns the best valid solution $(q_{i_\mathrm{val}}^*,\, \phi_{i_\mathrm{val}},\, d_{i_\mathrm{val}},\, i_\mathrm{val})$, whose nominal miscoverage level is computed via \cref{eq:nonconvexCovTradeoff} with $\neval=i$, as soon as one of two conditions holds: (i) the best valid solution would no longer be certifiable in the next iteration due to the shrinking per-iteration confidence budget, or (ii) the soft iteration limit $i_\mathrm{soft}$ is reached and a valid solution has been found. If no valid solution is found but all constraints are already satisfied, or if no valid solution is found within $i_\mathrm{max}$ iterations, the algorithm returns \texttt{None}, i.e., calibration fails.

\textbf{Practical considerations.}
The initial penalty $\phi_1$ can strongly influence convergence: a value that is too small may cause many iterations to be spent in the invalid regime, whereas a value that is too large may over-penalize violations and yield unnecessarily conservative prediction sets. In practice, $\phi_1$ can be tuned heuristically, e.g., by warm-starting from a related problem instance. The increase and decrease factors $f_+$ and $f_-$ control the speed of the search: larger $f_+$ and smaller $f_-$ lead to faster but coarser exploration, whereas values closer to $1$ yield finer but slower convergence. In our experiments (\cref{sec:experiments} and Appendix~\ref{sec:addResults}), \cref{alg:RelMCP} found a valid solution, i.e., one with $\varepsilon_i\leq\varepsilon$, in every run in which the target can be achieved at all; at small miscoverage levels, the target may not be achievable even without violations, e.g., for H-RelMCP-B at $\varepsilon\le0.02$ with $\ncal=2{,}000$. 

\textbf{Coverage guarantee.} 
\cref{thm:multiVarCP_nonconvex} provides a finite-sample, high-confidence coverage guarantee for RelMCP, i.e., for selecting a valid solution $\qe^*$ --- one satisfying $\varepsilon_i \leq \varepsilon$ --- from a {predetermined} grid of penalty parameters $\phi_1, \dots, \phi_{\neval}$, with $\varepsilon_i$ computed from the solution complexity or an upper bound thereof. H-RelMCP deviates from this setting in two respects. First, \cref{alg:RelMCP} constructs the grid {adaptively}: each penalty $\phi_i$ depends on the outcomes of all preceding solver calls, which themselves depend on the calibration data, whereas \cref{thm:multiVarCP_nonconvex} requires the grid to be fixed before calibration. Second, $d_i$ is approximated by the heuristic in Line 4, which is motivated by convex subproblems, such as those of H-RelMCP-A: at their unique optimum, the violated points and at most $\nq$ non-violated points determine the solution, so that the heuristic bounds the solution complexity of the exact minimizer from above. Our implementation, however, selects among several local solutions (Appendix~\ref{sec:experimentalDetails}), which need not coincide with this minimizer. If fewer non-violated points determine the solution, which becomes more likely for large $\nq$, the heuristic overestimates the solution complexity and the resulting certificate is conservative. For non-convex subproblems, such as those of H-RelMCP-B, however, the true solution complexity may be larger, so that the heuristic is not guaranteed to bound it from above. We therefore cannot establish a finite-sample guarantee for H-RelMCP, and its miscoverage levels $\varepsilon_i$ and the resulting misconfidence levels $\bar{\beta}$ are nominal values rather than certified bounds. 
Empirically, the adaptive search does not noticeably inflate the misconfidence. At $\beta=10^{-1}$, which admits up to $10\%$ of the runs below the target coverage (Appendix~\ref{subsec:varConfidence}), $9\%$ to $12\%$ of the SCP runs fall below it, but only $0.2\%$ of the H-RelMCP-A runs and none of the H-RelMCP-B runs; at smaller $\beta$, no H-RelMCP run does. A rigorous guarantee for the adaptive procedure is left for future work.

\section{Example Instantiations for the Score and Cost Functions}\label{sec:Examples}

We present representative score and cost function instantiations for regression tasks to illustrate the flexibility of MCP. Although the examples focus on regression, the framework extends straightforwardly to other settings such as classification or structured prediction~\citep{zhang2025confStructured}. Throughout, we assume that a point predictor $f(x)$ has been pretrained on a separate training dataset. The input-output pair $(x,y)$ may represent time series, $x=[x_1^\top~\cdots~x_{\nk}^\top]^\top$, $y=[y_1^\top~\cdots~y_{\nk}^\top]^\top$, with prediction sets $\Y = \Y_1 \times \cdots \times \Y_{\nk}$ over a time horizon $\nk$.

\subsection{Convex Prediction Set Shapes}\label{subsec:exConvex}
For convex prediction set families, the score function follows directly from the set membership condition.

\textbf{Intervals.} 
A natural and widely applicable choice is the multi-dimensional interval
\begin{align*}
    \Y(x,\qe)=\{ y \; | \; h(x,q_1) \leq y-f(x) \leq h(x,q_2) \}
\end{align*} 
with lower and upper bounds $h(x,\qe_1),h(x,\qe_2)\in \mathbb{R}^{\ny}$.
The corresponding score and cost functions are
\begin{subequations}
\begin{align}
    s(x,y,q) &=
\begin{bmatrix}
-y+f(x) + h(x,q_1) \\
\hphantom{-}y-f(x) - h(x,q_2)
\end{bmatrix}, \\
    J(\qe) &= \frac{1}{n_{\mathcal{X}}} \sum_{x\in\mathcal{X}} \| h(x,q_2) - h(x,q_1) \|,\label{eq:costInt}
\end{align}    
\end{subequations}
where $\mathcal{X}$ is a set of representative inputs (for example, derived from the training set $\Mtrain$) with cardinality $|\mathcal{X}|=n_{\mathcal{X}}$.
A sample $(x,y)$ is conformal if and only if $y$ lies within the interval; the cost penalizes total interval width across the training inputs.

\textbf{Ellipsoids.} 
MCP naturally accommodates ellipsoidal prediction sets of the form
\begin{align*}
    \Y(x,\qe) = \{ y \; | \; \bigl\|L(x,q)^\top (y-f(x) - h(x,q))\bigr\|_2 \leq 1 \}, 
\end{align*}
where the lower-triangular matrix $L(x,q)$ with positive diagonal entries determines the shape of the ellipsoid through its covariance matrix $\Sigma(x,q) = (L(x,q)L(x,q)^\top)^{-1}$, and $f(x) + h(x,q)$ is the ellipsoid center. The score and cost functions are
\begin{subequations}\label{eq:costScoreEllips}
\begin{align}
    s(x,y,\qe) &= \bigl\|L(x,q)^\top (y-f(x) - h(x,q))\bigr\|_2 - 1, ~\label{eq:scoreEllips}\\
J(\qe) &= -\frac{2}{n_{\mathcal{X}}} \sum_{x\in\mathcal{X}} \sum_{j=1}^{\ny} \log L(x,q)_{jj}, \label{eq:costEllips}
\end{align}    
\end{subequations}
where the cost equals $\frac{1}{n_{\mathcal{X}}} \sum_{x\in\mathcal{X}} \log \det \Sigma(x,q)$ and thus minimizes ellipsoid volume~\citep{bowman2023mvee}. When the ellipsoid shape is constant across inputs, $\qe$ collects the $\nq = \frac{1}{2}\ny(\ny+1)$ entries of the lower-triangular matrix $L$. If additionally $h$ does not depend on $\qe$, the score is convex in $\qe$ and the cost is convex for positive diagonal elements of $L$, making RemMCP directly applicable.

The squared form $\|L(x,q)^\top (y-f(x) - h(x,q))\|_2^2 - 1$ defines the same prediction sets and suffices for RemMCP, which only uses the sign of the score. For RelMCP, however, the norm without square, as in \cref{eq:scoreEllips}, makes the slack $\xi_m$ in \cref{eq:RelMCPprogram} grow linearly rather than quadratically with the distance of a calibration point from the prediction set, so that a few points far outside the set do not dominate the penalty. 

\textbf{Zonotopes.} 
Zonotopes are centrally symmetric polytopes well-suited to high-dimensional uncertainty quantification due to their compact parameterization and flexible geometry. A zonotopic prediction set takes the form
\begin{align*}
    \Y(x,\qe) =  \{ f(x) + c(x,q) + G(x,q) \lambda \; | \; \lambda\in [-\mathbf{1},\mathbf{1}] \}, 
\end{align*}
where $f(x) + c(x,q)\in \mathbb{R}^{\ny}$ is the zonotope center and $G(x,\qe)\in\mathbb{R}^{\ny\times n_{\mathrm{gen}}}$ is the generator matrix. The corresponding score and cost functions are
\begin{subequations}
\begin{align}
    s(x,y,\qe) &= 1-\mathbb{I}\{\exists \lambda\in [-\mathbf{1},\mathbf{1}]\colon y = f(x) + c(x,q) + G(x,q) \lambda \},\quad \\
J(\qe) &= \frac{1}{n_{\mathcal{X}}} \sum_{x\in\mathcal{X}} \sum_{i=1}^{n_{\mathrm{rot}}}\mathbf{1}^\top |R_i G(x,\qe)| \mathbf{1},\label{eq:costZon}    
\end{align}
\end{subequations}
where $R_i\in\mathbb{R}^{\ny\times\ny}$ are random rotation matrices and $|\cdot|$ denotes element-wise absolute value. The cost is a tractable proxy for zonotope volume~\citep{luetzow2024generator,luetzow2025zonoconformal}. When $c(x,\qe)$ is linear in $\qe$ and $G(x,\qe)=G_0(x)\,\mathrm{diag}(\qe_G)$ scales fixed generators by nonnegative entries $\qe_G$ of $\qe$~\citep{luetzow2025zonoconformal}, the calibration program \cref{eq:MCPprogram} reduces to a linear program via standard auxiliary variable reformulations.

\subsection{Sublevel Sets} \label{subsec:exNonconvex}

When the convex shapes above are insufficiently expressive, prediction sets can be defined as sublevel sets of arbitrary functions, enabling nonconvex geometries. Following \citet{dietrich2024nonconvex}, we construct the score function as a mixture of $N$ radial basis functions with centers $\mu_i\in\mathbb{R}^{\ny}$, widths $\sigma_i > 0$, and a threshold $\gamma\in(0,1)$ that is fixed before calibration (e.g., $\gamma=0.25$ in~\citet{dietrich2024nonconvex}):
\begin{subequations}\label{eq:costScoreRBF}
\begin{align}
    s(x,y,q) &= \gamma - \sum_{i=1}^N e^{-\frac{1}{2}\frac{\|y-f(x)-\mu_i\|_2^2}{\sigma_i^2}}, \\
    J(q) &= \sum_{i=1}^N \sigma_i^2,
\end{align}
\end{subequations}
with parameter vector $\qe = [\mu_1^\top~\cdots~\mu_N^\top~\sigma_1~\cdots~\sigma_N]^\top$. The cost penalizes the total spread of the basis functions, promoting compact prediction sets. 

\subsection{Unions of Prediction Sets} \label{subsec:exUnion}
Multi-modal prediction sets can be obtained by combining component score functions via the $\min$-operator, following the approach of~\citet{tumu2024MultiModalCP}, but without requiring an additional data split. 
Given partial score functions ${s}_{q_i}(x,y) \in \mathbb{R}^{\ns}$, $i=1,\dots,N$ --- each defined by any of the instantiations above --- and volume proxies $J_{q_i}$, the composite score and cost with $q=[q_1^\top~\dots~ q_N^\top]$ are 
\begin{subequations}\label{eq:MCPmultimod}
\begin{align}
{s}(x,y,\qe) 
&= \min_{i\in\{1,\dots,N\}} \max_{j\in\{1,\dots,\ns\}} \left(s_{q_i(j)}(x,y) \right), \\
J(q) &= \sum_{i=1}^N J_{q_i}.
\end{align}    
\end{subequations}
The resulting prediction set is the union of the $N$ component sets, with all shape parameters optimized and calibrated jointly over the full calibration dataset.

\section{Experimental Details}\label{sec:experimentalDetails}

This section provides additional details on the setups of the experiments in \cref{sec:experiments} and Appendices~\ref{sec:addResults} and~\ref{sec:expectedResults}.
All computations are carried out in Python on an AMD EPYC 9554 processor (3.1GHz).
Results are aggregated over 500 independent calibration runs. Prediction set volume is estimated via Monte Carlo sampling using $10^5$ uniform draws inside the smallest axis-aligned box that contains the prediction set, applied uniformly across all methods.

\subsection{Data}
\emph{Vehicle data.}
Each trajectory follows one of the three driving policies of~\citet{tumu2024MultiModalCP} (turning left, turning right, or going straight), drawn uniformly at random for each trajectory, so that the data points are i.i.d. As point predictor, we use the constant turn-rate and velocity model of~\citet{tumu2024MultiModalCP}, which requires no training: it extrapolates the trajectory from the positions, headings, and speeds observed over the past 5 time steps, using the last observed speed and the average turn rate over this window.
In each calibration run, we sample a new calibration set of $\ncal=2{,}000$ points and evaluate empirical coverage and prediction set volume on $n_\mathrm{test}=10{,}000$ newly sampled test points. For $T=2$, the residuals are scaled by $10^3$ in all figures, and the reported prediction set volumes refer to these scaled residuals, i.e., they are scaled by $10^6$.

\emph{CASP dataset.}
We predict the two-dimensional output (RMSD and F7) of the CASP regression benchmark~\citep{feldman2023multiQuantile}. The point predictor is a multilayer perceptron with two hidden layers of 64 ReLU neurons, trained on $5{,}000$ min-max normalized data points with Adam (learning rate $10^{-3}$) for at most 200 epochs, with early stopping on a 10\% validation split (patience of 20 epochs).
In each calibration run, we draw $\ncal=2{,}000$ calibration points and $n_\mathrm{test}=10{,}000$ test points without replacement from the remaining $40{,}730$ data points. Since the CASP residuals are unimodal, we only compare the methods with unimodal prediction sets (SCP-1, SCP-2A, SCP-3A, RemMCP, and H-RelMCP-A).

\emph{SCM20D dataset.}
We use the $7{,}463$ samples of the training split of the SCM20D supply-chain dataset~\citep{spyromitros2016rf1}, which has up to 16 outputs. The point predictor has the same architecture and training as for CASP, but is trained on $3{,}000$ randomly drawn data points for at most 500 epochs.
In each calibration run, we randomly split the remaining $4{,}463$ data points into $\ncal=2{,}000$ calibration points and $n_\mathrm{test}=2{,}463$ test points. As for CASP, the residuals are unimodal, so that we only compare the methods with unimodal prediction sets.

\subsection{Methods}
\emph{SCP-2.}
The calibration set is split into a fitting subset of size $n_\mathrm{cal} - n_\mathrm{cal}'$ and a calibration subset of size $\ncal' = \frac{1}{4}\ncal$. 
For SCP-2A, a single ellipsoid is fitted by computing the sample covariance $\hat{\Sigma}$ of the fitting residuals.
For SCP-2B, $k$-means clustering with $K=3$ is applied to the fitting residuals, and the sample covariance of each cluster is computed with a ridge $0.01\,\mathrm{tr}(\hat{\Sigma})/\ny$. The scores are normalized as in~\citet[Eq.~(9)]{tumu2024MultiModalCP}. Unlike~\citet{tumu2024MultiModalCP}, who cluster the high-density region of a kernel density estimate of the fitting residuals by mean shift, we apply $k$-means to the fitting residuals directly. On these clusters, the minimum-volume enclosing ellipsoids computed via CMA-ES, as in~\citet{tumu2024MultiModalCP}, resulted in higher variance and larger prediction sets than the sample covariance approach in our experiments; we therefore did not adopt them.
 
Unlike RemMCP and H-RelMCP, the split baselines use squared Mahalanobis distances as scores, to which the normalizing constants are applied for SCP-2B; since SCP only ranks the scores, the square does not affect SCP-2A. In both cases, the fitted shape is then held fixed and a single threshold $\tilde{q}^*$ is calibrated on the reserved split, with $\rho$ from \cref{eq:noutCP} (expected guarantee) or \cref{eq:noutMCPconf} (high-confidence guarantee) for $\nq=1$ and $\ncal'$ instead of $\ncal$ points.

\emph{SCP-3.} As SCP-2, but using $\ncal' = \frac{1}{2}\ncal$ data points for calibration.

\emph{RemMCP.}
The cascading scenario program is solved using SLSQP at each of the $\rho+1$ stages, with a maximum of $2{,}000$ iterations per stage.
 
\emph{H-RelMCP.}
Each penalized subproblem is solved with L-BFGS-B from $3$ (H-RelMCP-A) or $10$ (H-RelMCP-B) starts with at most $2{,}000$ or $1{,}000$ iterations each. The starts alternate between the best solution so far (for the best valid penalty or, if none exists yet, the best invalid one) and the initial guess, i.e., $L=I$ for H-RelMCP-A and a $k$-means fit over $\Mcal$ with per-cluster sample covariances for H-RelMCP-B; all starts except the first two are perturbed by Gaussian noise with standard deviation $0.25$ (H-RelMCP-A) or $0.5$ (H-RelMCP-B). Among the resulting candidates, we select the one with the smallest cost $J$ among those that violate at most as many calibration points as the candidate with the smallest penalized objective. %
In \cref{alg:RelMCP}, we set $i_\mathrm{soft}=100$, $i_\mathrm{max}=200$, $\phi_1=0.1$, $f_+=2$, and $f_-=0.3$.

\newpage
\section{Additional Experiments Targeting the High-Confidence Guarantee}\label{sec:addResults}

We present additional experimental results complementing those of \cref{sec:experiments}. All methods target the high-confidence guarantee in \cref{eq:confGuarantee}; we vary the miscoverage level $\varepsilon$, the misconfidence level $\beta$, and the output dimension $\ny$, each over 500 independent runs with randomly drawn calibration and test data. A method that cannot certify the target guarantee with the given number of calibration points is omitted; this applies when \cref{eq:noutMCPconf} admits no valid outlier budget or \cref{eq:nonconvexCovTradeoff} cannot certify the target guarantee even without violations.

\subsection{Varying Target Coverages} \label{subsec:varCoverage}

\begin{figure}[h!]
  \centering
  \sharedlegendSetsVehNew
  
\begin{subfigure}[t]{0.35\textwidth}
  \centering
    \includegraphics[page=1]{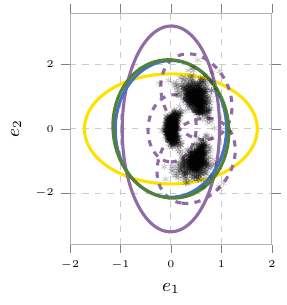}
  \caption{$\varepsilon=0.02$.}
\end{subfigure}%
\hfill
\begin{subfigure}[t]{0.32\textwidth}
  \centering
    \includegraphics[page=18]{figures/fig_environments/fig_sets_app.pdf}
  \caption{$\varepsilon=0.05$.}
\end{subfigure}%
\hfill
\begin{subfigure}[t]{0.32\textwidth}
  \centering
    \includegraphics[page=2]{figures/fig_environments/fig_sets_app.pdf}
  \caption{$\varepsilon=0.15$.}
\end{subfigure}
\vspace{-0.1cm}
\caption{\emph{Vehicle} prediction sets and calibration residuals ($*$) at $T=30$ and $\beta=10^{-6}$ for run 1. SCP-2A, SCP-2B, and H-RelMCP-B cannot certify the target guarantee at $\varepsilon=0.02$.}
  \label{fig:fig_vehicle_sets1}
\vspace{-0.1cm}
\end{figure}

\begin{figure}[h!]
  \centering
  \sharedlegendSetsVehSingle
  
\begin{subfigure}[t]{0.36\textwidth}
  \centering
    \includegraphics[page=3]{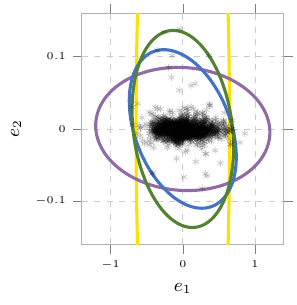}
  \caption{$\varepsilon=0.02$.}
\end{subfigure}%
\hfill
\begin{subfigure}[t]{0.317\textwidth}
  \centering
    \includegraphics[page=4]{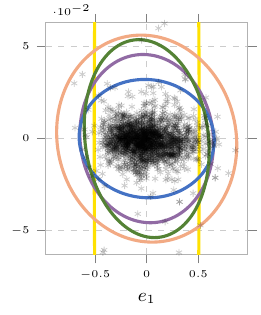}
  \caption{$\varepsilon=0.05$.}
\end{subfigure}%
\hfill
\begin{subfigure}[t]{0.319\textwidth}
  \centering
    \includegraphics[page=5]{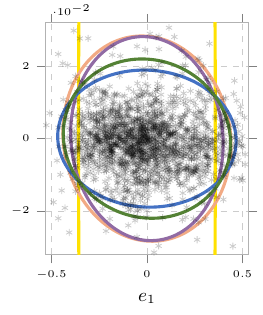}
  \caption{$\varepsilon=0.15$.}
\end{subfigure}
\vspace{-0.1cm}
\caption{\emph{CASP} prediction sets and calibration residuals ($*$) at $\beta=10^{-6}$ for run 1. SCP-2A cannot certify the target guarantee at $\varepsilon=0.02$; the SCP-1 balls exceed the plot range.}
  \label{fig:fig_CASP_sets}
\vspace{-0.1cm}
\end{figure}

We evaluate all methods across target miscoverage levels $\varepsilon \in \{0.02, 0.05, 0.15\}$ at $\beta=10^{-6}$ on the \emph{vehicle} prediction task, as in \cref{sec:experiments}, at $T=30$ time steps and on the \emph{CASP} dataset.
\Cref{fig:fig_vehicle_sets1} visualizes prediction sets and calibration residuals for \emph{vehicle} prediction, and \cref{fig:fig_CASP_sets} shows the corresponding results for the \emph{CASP} dataset. Box plots of empirical coverage and volume are reported in \cref{fig:fig_vehicle_box1,fig:fig_CASP_box}.

At $\varepsilon=0.02$, SCP-2, which reserves only $\ncal'=500$ points for threshold calibration, and H-RelMCP-B, which calibrates $\nq=15$ parameters, cannot certify the target guarantee.
All methods that can certify the target guarantee meet the target coverage $1-\varepsilon$ in every run. For \emph{vehicle} prediction, RemMCP yields the smallest single-ellipsoid sets at all miscoverage levels (mean volumes $7.54$, $5.64$, and $4.29$), closely followed by H-RelMCP-A ($7.85$, $5.75$, and $4.34$), whereas SCP-3A is even larger than SCP-1 at $\varepsilon=0.02$ ($9.97$ vs.\ $9.73$), since only $\rho=2$ of its $1{,}000$ threshold calibration points may be excluded. Among the multi-modal methods, SCP-3B yields the smallest sets at $\varepsilon=0.02$ ($5.48$; H-RelMCP-B cannot certify the target guarantee), while H-RelMCP-B yields the smallest sets at $\varepsilon=0.05$ ($2.76$ vs.\ $2.79$ for SCP-3B) and $\varepsilon=0.15$ ($1.33$ vs.\ $1.45$ for SCP-3B and $1.63$ for SCP-2B). For the \emph{CASP} dataset, RemMCP yields the smallest sets at all miscoverage levels (mean volumes $0.354$, $0.079$, and $0.029$). H-RelMCP-A yields smaller sets than the split SCP variants at all levels ($0.63$, $0.118$, and $0.034$ vs.\ $0.93$, $0.123$, and $0.053$ for SCP-3A), but larger ones than RemMCP at a comparable coverage (e.g., $0.977$ vs.\ $0.975$ at $\varepsilon=0.05$). This gap stems from the constraint relaxation: the penalty in \cref{eq:RelMCPprogram} grows with the distance of each violated calibration point from the set, so that the few points of the heavy-tailed \emph{CASP} residuals that lie far outside the set along $e_2$ pull the ellipsoid towards them, whereas RemMCP simply removes them.

Computation times and misconfidence levels (certified for SCP and RemMCP and nominal for H-RelMCP) are reported in \cref{tab:resultsVehicle30,tab:resultsVehicle30_conf}. The computation time of RemMCP grows with~$\varepsilon$ as the outlier budget $\rho$ increases, requiring more stages of the cascading program, whereas the cost of H-RelMCP is governed by the number of iterations of \cref{alg:RelMCP}.

\setcounter{topnumber}{5}
\setcounter{totalnumber}{10}

\begin{figure}[h!]
  \centering
  \sharedlegendVehNew
  
  \input{figures/rawTikz_app/fig_appHc_Vehicle30_epssweep_boxplot}
\vspace{-0.1cm}
\caption{Empirical coverage and prediction set volume for \emph{vehicle} data at $T=30$ and $\beta=10^{-6}$. %
Empty slots mark methods that cannot certify the target guarantee.}
  \label{fig:fig_vehicle_box1}
\vspace{-0.1cm}
\end{figure}

\begin{figure}[h!]
  \centering
  \sharedlegendVehSingle
  
  \input{figures/rawTikz_app/fig_appHc_CASP_epssweep_boxplot}
\vspace{-0.1cm}
\caption{Empirical coverage and prediction set volume for the \emph{CASP} dataset at $\beta=10^{-6}$. 
Empty slots mark methods that cannot certify the target guarantee.} %
  \label{fig:fig_CASP_box}
\vspace{-0.1cm}
\end{figure}

\begin{table}[h!]
\centering
\centering
\caption{Computation time [s] for \emph{vehicle} data at $T=30$ and $\beta=10^{-6}$ (mean $\pm$ standard deviation). A dash (--) indicates that the method cannot certify the target guarantee with $\ncal=2{,}000$.}
\label{tab:resultsVehicle30}
\footnotesize\setlength{\tabcolsep}{3pt}
\begin{tabular}{@{}lccc@{}}
\toprule
\textbf{Method} & $\varepsilon = 0.02$ & $\varepsilon = 0.05$ & $\varepsilon = 0.15$ \\
\midrule
SCP-1    & $0.0004 \pm 0.0000$ & $0.0005 \pm 0.0000$ & $0.0005 \pm 0.0000$ \\
SCP-2A   & -- & $0.0005 \pm 0.0000$ & $0.0005 \pm 0.0000$ \\
SCP-3A   & $0.0006 \pm 0.0000$ & $0.0005 \pm 0.0000$ & $0.0006 \pm 0.0000$ \\
RemMCP   & $0.0214 \pm 0.0022$ & $0.0776 \pm 0.0031$ & $0.2645 \pm 0.0042$ \\
H-RelMCP-A & $0.5779 \pm 0.3530$ & $0.4428 \pm 0.1701$ & $0.4174 \pm 0.0786$ \\
SCP-2B   & -- & $0.0054 \pm 0.0005$ & $0.0054 \pm 0.0005$ \\
SCP-3B   & $0.0049 \pm 0.0005$ & $0.0047 \pm 0.0001$ & $0.0047 \pm 0.0001$ \\
H-RelMCP-B & -- & $52.8 \pm 60.2$ & $35.9 \pm 24.3$ \\
\bottomrule
\end{tabular}
\end{table}

\begin{table}[h!]
\centering
\caption{Theoretical misconfidence $\bar{\beta}$ of achieving coverage $1-\varepsilon$ for \emph{vehicle} data at $T=30$ and $\beta=10^{-6}$, obtained from \cref{eq:noutMCPconf} and \cref{eq:nonconvexCovTradeoff} (mean).}
\label{tab:resultsVehicle30_conf}
\footnotesize\setlength{\tabcolsep}{3pt}
\begin{tabular}{@{}lccc@{}}
\toprule
\textbf{Method} & $\varepsilon = 0.02$ & $\varepsilon = 0.05$ & $\varepsilon = 0.15$ \\
\midrule
SCP-1    & $5.5{\cdot}10^{-7}$ & $6.9{\cdot}10^{-7}$ & $9.2{\cdot}10^{-7}$ \\
SCP-2A   & -- & $9.2{\cdot}10^{-7}$ & $7.8{\cdot}10^{-7}$ \\
SCP-3A   & $3.9{\cdot}10^{-7}$ & $7.7{\cdot}10^{-7}$ & $7.7{\cdot}10^{-7}$ \\
RemMCP   & $4.9{\cdot}10^{-8}$ & $6.9{\cdot}10^{-7}$ & $4.7{\cdot}10^{-7}$ \\
H-RelMCP-A & $9.6{\cdot}10^{-7}$ & $9.5{\cdot}10^{-7}$ & $9.4{\cdot}10^{-7}$ \\
SCP-2B   & -- & $9.2{\cdot}10^{-7}$ & $7.8{\cdot}10^{-7}$ \\
SCP-3B   & $3.9{\cdot}10^{-7}$ & $7.7{\cdot}10^{-7}$ & $7.7{\cdot}10^{-7}$ \\
H-RelMCP-B & -- & $9.2{\cdot}10^{-7}$ & $9.5{\cdot}10^{-7}$ \\
\bottomrule
\end{tabular}
\end{table}

\subsection{Varying Misconfidence Levels}\label{subsec:varConfidence}
We evaluate all methods for \emph{vehicle} position prediction at $T=30$ with $\varepsilon=0.05$ and $\beta\in\{10^{-9},10^{-6},10^{-3},10^{-1}\}$; the results for $\beta=10^{-6}$ are those of \cref{sec:experiments}. Box plots of empirical coverage and volume are shown in \cref{fig:fig_vehicle_box_beta}, and computation times and misconfidence levels in \cref{tab:resultsVehicle30_beta,tab:resultsVehicle30_beta_conf}.

The misconfidence $\bar{\beta}$ (nominal for H-RelMCP) does not exceed the target $\beta$ in any run (\cref{tab:resultsVehicle30_beta_conf}). Consistent with $\beta=10^{-1}$, between $9\%$ and $12\%$ of the runs of the SCP variants fall below the target coverage, compared with $3.6\%$ for RemMCP, $0.2\%$ for H-RelMCP-A, and none for H-RelMCP-B. %
A larger $\beta$ admits larger outlier budgets, which lowers coverage and set volume for all methods. The effect is strongest for the split SCP variants, whose outlier budget is limited by the few points reserved for threshold calibration (for SCP-2, $\rho$ increases from $1$ at $\beta=10^{-9}$ to $18$ at $\beta=10^{-1}$): from $\beta=10^{-9}$ to $\beta=10^{-1}$, the mean coverage of SCP-2 drops from $0.996$ to $0.962$, and the mean volume of SCP-2A shrinks from $9.74$, which exceeds that of SCP-1 ($8.03$), to $6.16$. In contrast, the mean volumes of RemMCP and H-RelMCP-A only shrink from $5.89$ to $5.20$ and from $5.94$ to $5.39$. Among the multi-modal methods, the mean volume of H-RelMCP-B shrinks from $3.01$ to $2.34$, that of SCP-3B from $3.07$ to $2.21$, and that of SCP-2B from $5.28$ to $2.29$, so that H-RelMCP-B yields the smallest sets for $\beta\leq10^{-6}$ but larger ones than SCP-3B at $\beta=10^{-1}$. Since RemMCP solves more stages of the cascading program for larger outlier budgets, its computation time increases with $\beta$ (\cref{tab:resultsVehicle30_beta}).

\begin{figure}[ht]
  \centering
  \sharedlegendVehNew
  
  \input{figures/rawTikz_app/fig_appHc_Vehicle30_betasweep_boxplot}
\vspace{-0.1cm}
\caption{Empirical coverage and prediction set volume for \emph{vehicle} data at $T=30$ and $\varepsilon=0.05$.} %
  \label{fig:fig_vehicle_box_beta}
\vspace{-0.1cm}
\end{figure}

\begin{table}[ht]
\centering
\caption{Computation time [s] for \emph{vehicle} position prediction at $T=30$ and $\varepsilon=0.05$ (mean $\pm$ standard deviation).}
\label{tab:resultsVehicle30_beta}
\footnotesize\setlength{\tabcolsep}{3.5pt}
\begin{tabular}{@{}lcccc@{}}
\toprule
\textbf{Method} & $\beta = 10^{-9}$ & $\beta = 10^{-6}$ & $\beta = 10^{-3}$ & $\beta = 10^{-1}$ \\
\midrule
SCP-1    & $0.0005 \pm 0.0000$ & $0.0005 \pm 0.0000$ & $0.0005 \pm 0.0000$ & $0.0005 \pm 0.0000$ \\
SCP-2A   & $0.0005 \pm 0.0001$ & $0.0005 \pm 0.0000$ & $0.0005 \pm 0.0000$ & $0.0005 \pm 0.0000$ \\
SCP-3A   & $0.0005 \pm 0.0000$ & $0.0005 \pm 0.0000$ & $0.0005 \pm 0.0000$ & $0.0005 \pm 0.0000$ \\
RemMCP   & $0.0632 \pm 0.0028$ & $0.0776 \pm 0.0031$ & $0.0913 \pm 0.0032$ & $0.1129 \pm 0.0034$ \\
H-RelMCP-A & $0.4501 \pm 0.2106$ & $0.4428 \pm 0.1701$ & $0.2733 \pm 0.1162$ & $0.3425 \pm 0.0754$ \\
SCP-2B   & $0.0054 \pm 0.0006$ & $0.0054 \pm 0.0005$ & $0.0054 \pm 0.0005$ & $0.0054 \pm 0.0005$ \\
SCP-3B   & $0.0047 \pm 0.0001$ & $0.0047 \pm 0.0001$ & $0.0047 \pm 0.0001$ & $0.0047 \pm 0.0001$ \\
H-RelMCP-B & $55.9 \pm 71.3$ & $52.8 \pm 60.2$ & $31.4 \pm 32.9$ & $25.9 \pm 23.0$ \\
\bottomrule
\end{tabular}%
\end{table}

\begin{table}[ht]
\centering
\caption{Misconfidence $\bar{\beta}$ of achieving coverage $1-\varepsilon$ for \emph{vehicle} position prediction at $T=30$ and $\varepsilon=0.05$, obtained from \cref{eq:noutMCPconf} and \cref{eq:nonconvexCovTradeoff} (mean).}
\label{tab:resultsVehicle30_beta_conf}
\footnotesize\setlength{\tabcolsep}{3.5pt}
\begin{tabular}{@{}lcccc@{}}
\toprule
\textbf{Method} & $\beta = 10^{-9}$ & $\beta = 10^{-6}$ & $\beta = 10^{-3}$ & $\beta = 10^{-1}$ \\
\midrule
SCP-1    & $5.7{\cdot}10^{-10}$ & $6.9{\cdot}10^{-7}$ & $7.6{\cdot}10^{-4}$ & $0.0979$ \\
SCP-2A   & $2.0{\cdot}10^{-10}$ & $9.2{\cdot}10^{-7}$ & $4.6{\cdot}10^{-4}$ & $0.0865$ \\
SCP-3A   & $9.4{\cdot}10^{-10}$ & $7.7{\cdot}10^{-7}$ & $7.2{\cdot}10^{-4}$ & $0.0806$ \\
RemMCP   & $1.1{\cdot}10^{-10}$ & $6.9{\cdot}10^{-7}$ & $3.3{\cdot}10^{-4}$ & $0.0807$ \\
H-RelMCP-A & $9.6{\cdot}10^{-10}$ & $9.5{\cdot}10^{-7}$ & $9.0{\cdot}10^{-4}$ & $0.0927$ \\
SCP-2B   & $2.0{\cdot}10^{-10}$ & $9.2{\cdot}10^{-7}$ & $4.6{\cdot}10^{-4}$ & $0.0865$ \\
SCP-3B   & $9.4{\cdot}10^{-10}$ & $7.7{\cdot}10^{-7}$ & $7.2{\cdot}10^{-4}$ & $0.0806$ \\
H-RelMCP-B & $9.2{\cdot}10^{-10}$ & $9.2{\cdot}10^{-7}$ & $9.1{\cdot}10^{-4}$ & $0.0936$ \\
\bottomrule
\end{tabular}%
\end{table}

\FloatBarrier

\subsection{Varying Output Dimensions}\label{subsec:varDim}

\begin{figure}[ht]
  \centering
  \sharedlegendSetsVehSingle
  
\begin{subfigure}[t]{0.27\textwidth}
  \centering
    \includegraphics[page=6]{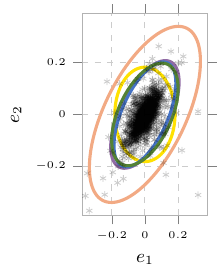}
  \caption{$\ny=2$.}
\end{subfigure}%
\hfill
\begin{subfigure}[t]{0.24\textwidth}
  \centering
    \includegraphics[page=7]{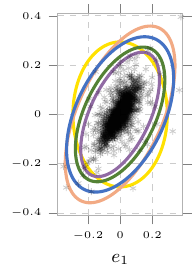}
  \caption{$\ny=4$.}
\end{subfigure}%
\hfill
\begin{subfigure}[t]{0.24\textwidth}
  \centering
    \includegraphics[page=8]{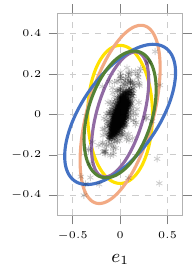}
  \caption{$\ny=6$.}
\end{subfigure}%
\hfill
\begin{subfigure}[t]{0.24\textwidth}
  \centering
    \includegraphics[page=9]{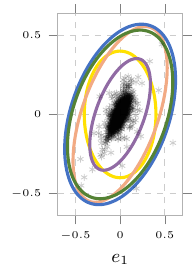}
  \caption{$\ny=8$.}
\end{subfigure}
\vspace{-0.1cm}
\caption{\emph{SCM20D} prediction sets and calibration residuals ($*$) at $\varepsilon=0.05$ and $\beta=10^{-6}$ for run 1, visualized over the first two output dimensions.}
  \label{fig:fig_SCM20D_sets}
\vspace{-0.1cm}
\end{figure}

We evaluate the effect of increasing output dimension on the \emph{SCM20D} dataset~\citep{spyromitros2016rf1}, considering output dimensions $\ny\in\{2,4,6,8\}$, while calibrating at $\varepsilon=0.05$ and $\beta=10^{-6}$. We do not consider $\ny \geq 10$: the ellipsoid covariance parameterization yields $\nq = \frac{1}{2}\ny(\ny+1) = 55$ free parameters at $\ny=10$, for which H-RelMCP-A cannot certify the target guarantee with $\ncal=2{,}000$ (it would need at least $2{,}121$ points) and \cref{eq:noutMCPconf} only admits the outlier budget $\rho=0$ for RemMCP. Prediction sets over the first two output dimensions and box plots of empirical coverage and volume are shown in \cref{fig:fig_SCM20D_sets,fig:fig_SCM20D_box}; computation times are in \cref{tab:resultsSCM20D}.

At $\ny=2$, RemMCP yields the smallest prediction sets (mean volume $0.091$ vs.\ $0.101$ for H-RelMCP-A, $0.109$ for SCP-1, and $0.111$ for SCP-3A). For larger $\ny$, RemMCP becomes increasingly conservative: its mean coverage rises to $0.982$, $0.988$, and $0.992$ at $\ny=4$, $6$, and $8$, and its certified misconfidence falls far below the target ($5.9\cdot10^{-9}$, $7.5\cdot10^{-12}$, and $1.8\cdot10^{-14}$).
The main reason is that the calibration problems are far from fully supported (\cref{assumption:fullSupport}): under full support, the outlier budgets $\rho=4$, $1$, and $0$ would yield mean coverages of only $0.975$, $0.979$, and $0.982$ by \cref{eq:betaMCP}, whereas at $\ny=2$, the observed and predicted coverages agree ($0.973$ vs.\ $0.972$). Accordingly, the share of the $\rho\nq$ removed calibration points that lie outside the final prediction set drops from $92\%$ at $\ny=2$ to $68\%$ and $53\%$ at $\ny=4$ and $6$, i.e., the support sets contain fewer than $\nq$ points and are filled up with non-violated samples.
H-RelMCP-A, whose nominal a posteriori certificate counts the violated points, is less conservative (e.g., mean volume $0.019$ vs.\ $0.051$ at $\ny=6$) but also exceeds the target coverage considerably at $\ny=8$ ($0.990$). For $\ny\geq4$, SCP-3A yields the smallest sets, closely followed by H-RelMCP-A at $\ny=4$ ($0.0245$ vs.\ $0.0249$). Thus, for calibration problems that are far from being fully supported, MCP is expected to yield more conservative prediction sets.

\begin{figure}[ht]
  \centering
  \sharedlegendVehSingle
  
  \input{figures/rawTikz_app/fig_appHc_SCM20D_dimsweep_boxplot}
\vspace{-0.1cm}
\caption{Empirical coverage and prediction set volume for the \emph{SCM20D} dataset at $\varepsilon=0.05$ and $\beta=10^{-6}$ over $2{,}463$ test points and 500 calibration runs.}
  \label{fig:fig_SCM20D_box}
\vspace{-0.1cm}
\end{figure}

\begin{table}[ht]
\centering
\caption{Computation time [s] for the \emph{SCM20D} dataset at $\varepsilon=0.05$ and $\beta=10^{-6}$ (mean $\pm$ standard deviation).}
\label{tab:resultsSCM20D}
\footnotesize\setlength{\tabcolsep}{3.5pt}
\begin{tabular}{@{}lcccc@{}}
\toprule
\textbf{Method} & $\ny=2$ & $\ny=4$ & $\ny=6$ & $\ny=8$ \\
\midrule
SCP-1    & $0.0004 \pm 0.0000$ & $0.0004 \pm 0.0000$ & $0.0005 \pm 0.0000$ & $0.0005 \pm 0.0000$ \\
SCP-2A   & $0.0005 \pm 0.0000$ & $0.0006 \pm 0.0000$ & $0.0006 \pm 0.0000$ & $0.0007 \pm 0.0000$ \\
SCP-3A   & $0.0005 \pm 0.0000$ & $0.0006 \pm 0.0000$ & $0.0006 \pm 0.0000$ & $0.0007 \pm 0.0000$ \\
RemMCP   & $0.0858 \pm 0.0028$ & $0.1098 \pm 0.0081$ & $0.0926 \pm 0.0090$ & $0.0862 \pm 0.0067$ \\
H-RelMCP-A & $0.3304 \pm 0.1281$ & $0.4915 \pm 0.3171$ & $0.9662 \pm 0.8051$ & $7.5642 \pm 6.8870$ \\
\bottomrule
\end{tabular}%
\end{table}

\FloatBarrier
\quad 

\newpage   %

\section{Experiments Targeting the Expected Coverage Guarantee}\label{sec:expectedResults}

This section reports the \emph{vehicle} prediction experiments of \cref{sec:experiments} and the miscoverage sweeps of Appendix~\ref{subsec:varCoverage} with the expected guarantee in \cref{eq:expGuarantee} as the target. The SCP baselines and RemMCP (with $\rho$ from \cref{eq:noutMCP}) are configured as in \cref{sec:experiments}. H-RelMCP is omitted, since it only targets the high-confidence guarantee.

\subsection{Varying Prediction Horizons}\label{subsec:expHorizons}

\begin{figure}[h!]
  \centering
  \sharedlegendSetsVehNoRel
  \vspace{-0.1cm}
  
\begin{subfigure}[t]{0.35\textwidth}
  \centering
    \includegraphics[page=10, trim=0cm 0.19cm 0cm 0cm, clip]{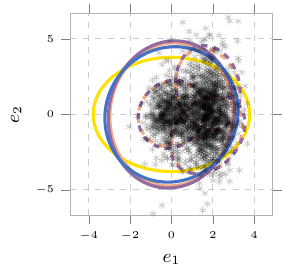}
  \caption{$T=2$ (residuals scaled by $10^3$).}
  \label{fig:exp_vehicleT2_sets}
\end{subfigure}%
\hfill
\begin{subfigure}[t]{0.32\textwidth}
  \centering
    \includegraphics[page=11, trim=0cm 0.19cm 0cm 0cm, clip]{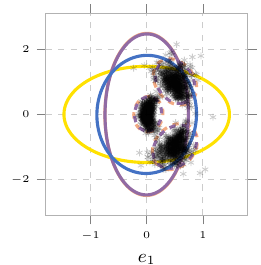}
  \caption{$T=30$.}
  \label{fig:exp_vehicle30_sets}
\end{subfigure}%
\hfill
\begin{subfigure}[t]{0.325\textwidth}
  \centering
    \includegraphics[page=12, trim=0cm 0.19cm 0cm 0cm, clip]{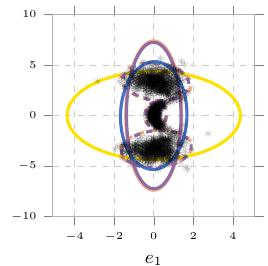}
  \caption{$T=50$.}
  \label{fig:exp_vehicle50_sets}
\end{subfigure}
\vspace{-0.1cm}
\caption{\emph{Vehicle} prediction sets and calibration residuals ($*$) at $\alpha=0.05$ under the expected guarantee for run 1.}
  \label{fig:exp_vehicle_sets}
\vspace{-0.1cm}
\end{figure}

We repeat the \emph{vehicle} prediction experiments of \cref{sec:experiments} at $T=2$, $30$, and $50$ with $\alpha=0.05$.
\Cref{fig:exp_vehicle_sets} visualizes prediction sets and calibration residuals and \cref{fig:exp_vehicle_box} reports empirical coverage and prediction set volume across 500 independent calibration runs. The SCP variants attain the target coverage on average (mean coverage $0.950$), whereas RemMCP slightly exceeds it ($0.952$--$0.954$), since its coverage bound is only tight under \cref{assumption:fullSupport,assumption:scenarioViolation}.
Among single-ellipsoid methods, RemMCP yields the tightest sets at $T=30$ and $T=50$ and is on par with SCP-1 at $T=2$ (mean volumes $46.5$, $5.08$, and $28.0$ at $T=2$, $30$, and $50$, compared with $46.4$, $6.96$, and $60.1$ for SCP-1 and $48.3$, $5.85$, and $29.6$ for SCP-3A). The multi-modal methods SCP-2B and SCP-3B yield comparable sets at $T=30$ ($2.04$ and $2.03$) and $T=50$ ($21.9$ for both), whereas at $T=2$, a few degenerate runs ($32$ and $29$ of $500$) inflate their mean volumes to $797$ and $449$.
All SCP-2 and SCP-3 variants exhibit a higher variance in coverage, a direct consequence of the data split reducing the number of points available for threshold calibration. For $T=30$, the standard deviation in coverage ranges from $0.0072$ to $0.0098$ for SCP-2 and SCP-3 variants, compared with $0.0054$ for SCP-1 and $0.0055$ for RemMCP.
The coverage distributions in \cref{fig:exp_betaDistr} explain these differences. Under the expected guarantee, the distributions of all methods have their mean at the target, but a data split widens them.
Regarding computation, SCP-1, SCP-2, and SCP-3 complete in under $10\,\mathrm{ms}$, and RemMCP requires approximately $0.13\,\mathrm{s}$.

\begin{figure}[h!]
  \centering
  \sharedlegendVehNoRel
  
  \input{figures/rawTikz_app/fig_appExpect_Vehicle_boxplot}
\vspace{-0.1cm}
\caption{Empirical coverage and prediction set volume for \emph{vehicle} data at $\alpha=0.05$ under the expected guarantee.}
  \label{fig:exp_vehicle_box}
\vspace{-0.1cm}
\end{figure}

\begin{figure}[h!]
  \centering
  \begin{tikzpicture}
\begin{axis}[
    width=6.5cm,
    height=5cm,
    xlabel={Coverage Probability $\eta(q)$},
    ylabel={},
    xmin=0.9, xmax=1.0,
    ymin=0,
    ytick=\empty,
    xtick={0.9,0.95,1.0},
    xticklabel style={font=\small},
    xlabel style={font=\small},
    ylabel style={font=\small},
    axis lines=left,
    axis line style={-},
    tick style={color=black, thin},
    clip=false,
    legend style={
        at={(0.,0.97)},
        anchor=north west,
        font=\footnotesize,
        draw=none,
        fill=none,
        row sep=1pt,
    },
    legend cell align=left,
]

\addplot[
    color=scp1color,
    line width=2pt,restrict x to domain=0.9:inf
] table {figures/fig_helpers/beta_coords_ncal2000_eps5_nq1_rho99};
\addlegendentry{SCP-1} %

\addplot[
    color=scp3color,
    line width=1.4pt,restrict x to domain=0.9:inf
] table {figures/fig_helpers/beta_coords_ncal500_eps5_nq1_rho24};
\addlegendentry{{SCP-2}} %

\addplot[
    color=scp4color,
    line width=1.4pt,restrict x to domain=0.9:inf
] table {figures/fig_helpers/beta_coords_ncal1000_eps5_nq1_rho49};
\addlegendentry{{SCP-3}} %

\addplot[
    color=cmcp1color,
    line width=1.2pt,restrict x to domain=0.9:inf
] table {figures/fig_helpers/beta_coords_ncal2000_eps5_nq3_rho32};
\addlegendentry{RemMCP} %

\addplot[
    color=black!40!red,
    dashed,
    line width=1.0pt,
] coordinates {(0.95, 0) (0.95, 90)};
\node[
    anchor=south west,
    font=\footnotesize,
    text=black!30!red,
] at (axis cs: 0.95, 77) {~$1-\alpha$};
\end{axis}
\end{tikzpicture}
  \caption{Coverage distributions for $\alpha=0.05$ from \cref{eq:betaMCP} with $\rho$ from \cref{eq:noutMCP}. For RemMCP, the coverage is in general stochastically larger, with equality under \cref{assumption:fullSupport,assumption:scenarioViolation}.}
  \label{fig:exp_betaDistr}
\end{figure}

\subsection{Varying Target Coverages} \label{subsec:expVarCoverage}

\begin{figure}[h!]
  \centering
  \sharedlegendSetsVehNoRel
  
\begin{subfigure}[t]{0.35\textwidth}
  \centering
    \includegraphics[page=13]{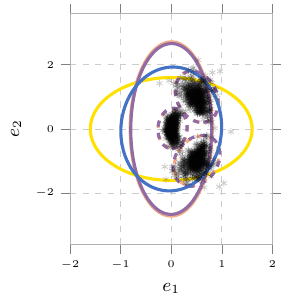}
  \caption{$\alpha=0.02$.}
\end{subfigure}%
\hfill
\begin{subfigure}[t]{0.32\textwidth}
  \centering
    \includegraphics[page=19]{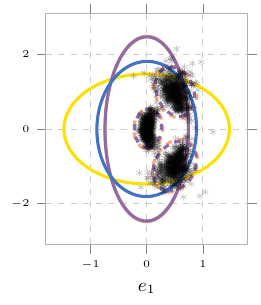}
  \caption{$\alpha=0.05$.}
\end{subfigure}%
\hfill
\begin{subfigure}[t]{0.32\textwidth}
  \centering
    \includegraphics[page=14]{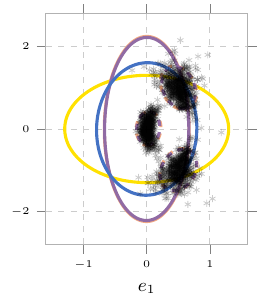}
  \caption{$\alpha=0.15$.}
\end{subfigure}
\vspace{-0.1cm}
\caption{\emph{Vehicle} prediction sets and calibration residuals ($*$) at $T=30$ under the expected guarantee for run 1.}
  \label{fig:exp_vehicle_sets1}
\vspace{-0.1cm}
\end{figure}

\begin{figure}[h!]
  \centering
  \sharedlegendSetsVehSingleNoRel
  
\begin{subfigure}[t]{0.36\textwidth}
  \centering
    \includegraphics[page=15]{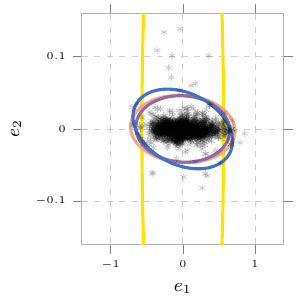}
  \caption{$\alpha=0.02$.}
\end{subfigure}%
\hfill
\begin{subfigure}[t]{0.317\textwidth}
  \centering
    \includegraphics[page=16]{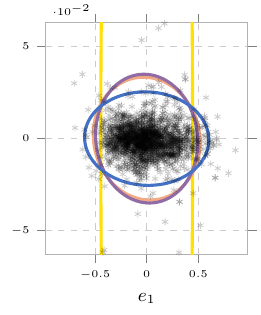}
  \caption{$\alpha=0.05$.}
\end{subfigure}%
\hfill
\begin{subfigure}[t]{0.319\textwidth}
  \centering
    \includegraphics[page=17]{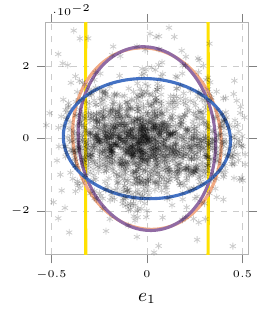}
  \caption{$\alpha=0.15$.}
\end{subfigure}
\vspace{-0.1cm}
\caption{\emph{CASP} prediction sets and calibration residuals ($*$) under the expected guarantee for run 1. The SCP-1 balls exceed the plot range.}
  \label{fig:exp_CASP_sets}
\vspace{-0.1cm}
\end{figure}

We evaluate all methods under the expected guarantee across target miscoverage levels $\alpha \in \{0.02, 0.05, 0.15\}$ for \emph{vehicle} prediction at $T=30$ and for the \emph{CASP} dataset, as described in Appendix~\ref{subsec:varCoverage}. Prediction sets are shown in \cref{fig:exp_vehicle_sets1,fig:exp_CASP_sets}, box plots of empirical coverage and volume in \cref{fig:exp_vehicle_box1,fig:exp_CASP_box}, and computation times and guaranteed confidence levels in \cref{tab:expVehicle30,tab:expVehicle30_conf}.
The SCP variants attain a mean coverage within $0.002$ of the target at all miscoverage levels, and RemMCP slightly exceeds it (by $0.003$--$0.004$). For \emph{vehicle} prediction, RemMCP yields the smallest single-ellipsoid sets at all levels (mean volumes $6.04$, $5.08$, and $4.02$ vs.\ $6.87$, $5.85$, and $4.69$ for SCP-3A). For the \emph{CASP} dataset, RemMCP yields the smallest sets at all levels (mean volumes $0.107$, $0.052$, and $0.024$ vs.\ $0.126$, $0.079$, and $0.041$ for SCP-3A). As $\alpha$ increases, RemMCP solves more stages of the cascading program, so that its computation time grows from $0.06\,\mathrm{s}$ to $0.34\,\mathrm{s}$ (\cref{tab:expVehicle30}). Finally, \cref{tab:expVehicle30_conf} shows that the expected guarantee only certifies a confidence of $0.51$ to $0.59$ that the coverage of a given calibration run reaches the target; indeed, $43\%$ to $51\%$ of the SCP runs and $17\%$ to $31\%$ of the RemMCP runs fall below $1-\alpha$ for both datasets. Applications that require the target coverage for the calibration data at hand thus call for the high-confidence guarantee targeted in \cref{sec:experiments} and Appendix~\ref{sec:addResults}.

\begin{figure}[h!]
  \centering
  \sharedlegendVehNoRel
  
  \input{figures/rawTikz_app/fig_appExpect_Vehicle30_epssweep_boxplot}
\vspace{-0.1cm}
\caption{Empirical coverage and prediction set volume for \emph{vehicle} data at $T=30$ under the expected guarantee.}%
  \label{fig:exp_vehicle_box1}
\vspace{-0.1cm}
\end{figure}

\begin{figure}[h!]
  \centering
  \sharedlegendVehSingleNoRel
  
  \input{figures/rawTikz_app/fig_appExpect_CASP_epssweep_boxplot}
\vspace{-0.1cm}
\caption{Empirical coverage and prediction set volume for the \emph{CASP} dataset under the expected guarantee.} %
  \label{fig:exp_CASP_box}
\vspace{-0.1cm}
\end{figure}
\begin{table}[h!]
\centering
\centering
\caption{Computation time [s] for \emph{vehicle} data at $T=30$ under the expected guarantee (mean $\pm$ standard deviation).}
\label{tab:expVehicle30}
\footnotesize\setlength{\tabcolsep}{3.5pt}
\begin{tabular}{@{}lccc@{}}
\toprule
\textbf{Method} & $\alpha = 0.02$ & $\alpha = 0.05$ & $\alpha = 0.15$ \\
\midrule
SCP-1    & $0.0001 \pm 0.0000$ & $0.0001 \pm 0.0000$ & $0.0001 \pm 0.0000$ \\
SCP-2A   & $0.0004 \pm 0.0000$ & $0.0004 \pm 0.0000$ & $0.0004 \pm 0.0000$ \\
SCP-3A   & $0.0003 \pm 0.0000$ & $0.0003 \pm 0.0000$ & $0.0003 \pm 0.0000$ \\
RemMCP   & $0.0560 \pm 0.0027$ & $0.1262 \pm 0.0035$ & $0.3422 \pm 0.0045$ \\
SCP-2B   & $0.0053 \pm 0.0005$ & $0.0053 \pm 0.0005$ & $0.0052 \pm 0.0005$ \\
SCP-3B   & $0.0045 \pm 0.0001$ & $0.0045 \pm 0.0001$ & $0.0044 \pm 0.0001$ \\
\bottomrule
\end{tabular}
\end{table}

\begin{table}[h!]
\centering
\caption{Guaranteed confidence $1-\bar{\beta}$ of achieving coverage $1-\alpha$ for \emph{vehicle} data at $T=30$ under the expected guarantee, obtained from \cref{eq:noutMCPconf} (mean).}
\label{tab:expVehicle30_conf}
\footnotesize\setlength{\tabcolsep}{3.5pt}
\begin{tabular}{@{}lccc@{}}
\toprule
\textbf{Method} & $\alpha = 0.02$ & $\alpha = 0.05$ & $\alpha = 0.15$ \\
\midrule
SCP-1    & $0.5217$ & $0.5143$ & $0.5096$ \\
SCP-2A   & $0.5433$ & $0.5286$ & $0.5191$ \\
SCP-3A   & $0.5306$ & $0.5203$ & $0.5135$ \\
RemMCP   & $0.5852$ & $0.5552$ & $0.5096$ \\
SCP-2B   & $0.5433$ & $0.5286$ & $0.5191$ \\
SCP-3B   & $0.5306$ & $0.5203$ & $0.5135$ \\
\bottomrule
\end{tabular}
\end{table}

\newpage

\section{Comparison of RemMCP, RelMCP, and H-RelMCP}\label{sec:comparison}

This section compares RemMCP, RelMCP, and its heuristic version H-RelMCP with respect to flexibility, optimality, coverage guarantees, and computational complexity; \cref{tab:comparisonMCP} summarizes the main differences. In short, RemMCP provides a direct generalization of split CP but is restricted to convex formulations, whereas RelMCP offers high flexibility at the expense of higher computational cost; H-RelMCP, which we use in the experiments, avoids predefining a grid of penalty parameters at the price of a missing guarantee.

\begin{table}[ht]
\centering
\caption{Comparison of RemMCP, RelMCP, and H-RelMCP.}
\label{tab:comparisonMCP}
\small\setlength{\tabcolsep}{4pt}
\begin{tabular}{@{}>{\raggedright\arraybackslash}p{0.15\textwidth}>{\raggedright\arraybackslash}p{0.26\textwidth}>{\raggedright\arraybackslash}p{0.26\textwidth}>{\raggedright\arraybackslash}p{0.26\textwidth}@{}}
\toprule
 & \textbf{RemMCP} & \textbf{RelMCP} & \textbf{H-RelMCP} \\
\midrule
\textbf{Score and cost functions} & Convex $J$ and convex sets $\{\qe \mid s\leq\mathbf{0}\}$; restricted set shapes & Possibly non-convex; highly flexible set shapes & As RelMCP \\
\addlinespace
\textbf{Subproblems} & $\rho+1$ constrained convex programs, with $\rho$ from \cref{eq:noutMCP} or \cref{eq:noutMCPconf} & $\neval$ unconstrained, possibly non-convex programs on a predetermined grid of penalty parameters & Unconstrained, possibly non-convex programs; penalties and their number (at most $i_\mathrm{max}$) set adaptively by \cref{alg:RelMCP} \\
\addlinespace
\textbf{Optimality} & Globally optimal solutions & No guarantee of optimality & No guarantee of optimality \\
\addlinespace
\textbf{Coverage guarantees} & Expected and high-confidence (\cref{thm:multiVarCP}); a priori; exact under \cref{assumption:fullSupport,assumption:scenarioViolation} & High-confidence (\cref{thm:multiVarCP_nonconvex}); a posteriori & None proven; nominal high-confidence certificate with heuristic $d_i$ \\
\addlinespace
\textbf{Empirical behavior} & Consistent with \cref{thm:multiVarCP} & Not evaluated in our experiments & Share of runs below target coverage at most $\beta$; far below it at $\beta=10^{-1}$ \\
\bottomrule
\end{tabular}
\end{table}

\textbf{Flexibility and optimality.}
RemMCP requires convex cost functions and score functions with convex feasible parameter sets (\cref{assumption:convex}), which restricts the prediction set shapes to, e.g., hyperrectangles, ellipsoids, and zonotopes (Appendix~\ref{subsec:exConvex}). In return, each program in \cref{eq:MCPprogram} is convex and can be solved to global optimality at comparably low computational cost. RelMCP and H-RelMCP admit arbitrary non-convex score and cost functions, enabling highly flexible shapes such as unions of ellipsoids with jointly optimized centers and covariances, which yield substantially smaller prediction sets for multi-modal residuals (H-RelMCP-B in \cref{sec:experiments}). However, the programs in \cref{eq:RelMCPprogram} are non-convex in general, so that reaching the optimal solution is not guaranteed. %

\textbf{Coverage guarantees.}
RemMCP provides a priori guarantees whose assumptions are straightforward to satisfy (\cref{subsec:scenario}); under \cref{assumption:fullSupport,assumption:scenarioViolation}, the coverage distribution is even characterized exactly by \cref{eq:betaMCP}. RelMCP only provides the high-confidence guarantee in \cref{eq:confGuarantee}, which is evaluated a posteriori. It requires a predetermined grid of penalty parameters and the solution complexity $d_i$ or an upper bound thereof (\cref{thm:multiVarCP_nonconvex}). H-RelMCP relaxes both requirements by choosing the penalties adaptively and approximating $d_i$ heuristically (Appendix~\ref{sec:RelMCPalg}). It therefore carries no proven finite-sample guarantee, although its share of runs below the target coverage never exceeds $\beta$ in our experiments and stays far below it at $\beta=10^{-1}$ (Appendix~\ref{subsec:varConfidence}).

\textbf{Computational complexity.}
Since MCP is agnostic to the optimizer, whose per-solve cost dominates the computation time and is chosen by the user, we characterize the complexity by the number of subproblems and how each of them scales. RemMCP solves $\rho+1$ convex programs, which equals $\bigl\lfloor \alpha(\ncal+1)/\nq \bigr\rfloor$ for $\rho$ from \cref{eq:noutMCP}; this number increases with $\alpha\ncal$ and decreases with $\nq$, and for $\rho$ from \cref{eq:noutMCPconf}, it additionally increases with $\beta$. Each program has $\nq$ optimization variables and a number of constraints that grows with $\ncal$, so that the cost per program grows with both $\nq$ and $\ncal$. This matches \cref{tab:resultsVehicle30}, where the computation time increases with $\varepsilon$, \cref{tab:resultsVehicle30_beta}, where it increases with $\beta$, and \cref{tab:resultsSCM20D}, where it stays between $0.08\,\mathrm{s}$ and $0.11\,\mathrm{s}$ as $\nq$ grows from $3$ to $36$, even though the number of programs decreases from $19$ to $1$. RelMCP solves $\neval$ unconstrained, possibly non-convex programs over $\nq$ optimization variables with $\ncal$ penalty terms and evaluates \cref{eq:nonconvexCovTradeoff} for each of them. For H-RelMCP, the number of programs equals the number of iterations of \cref{alg:RelMCP}, which depends on the initialization and the user-defined algorithm settings and is thus not known in advance, but bounded by $i_\mathrm{max}$.

\section{Limitations and Future Work} \label{sec:lim}

While MCP offers a principled and flexible framework for joint prediction set design and calibration, several limitations merit discussion.

The most immediate practical limitation is computational: by jointly optimizing shape and threshold over the full calibration set, MCP incurs substantially higher calibration cost than SCP, which calibrates a single scalar threshold in closed form. This cost grows with the number of optimization variables $\nq$ and the size of the calibration set $\ncal$, and is most pronounced for RelMCP and H-RelMCP, which require repeated solving of penalized subproblems. In settings where calibration time is critical, SCP remains the more practical choice.

A related limitation concerns data requirements. SCP baselines can leverage the training dataset for shape parameter estimation. MCP, by contrast, relies entirely on the calibration set for both stages simultaneously. While this is precisely what eliminates the variance-inflating data split, it also means that MCP requires a sufficiently large calibration set to certify coverage --- a requirement that becomes more demanding as $\nq$ grows.

Within the two proposed variants, RemMCP is restricted to convex cost functions and score functions with convex feasible parameter sets, which limits the prediction set shapes it can directly optimize. While convexity covers a broad and practically relevant class of shapes --- including ellipsoids, zonotopes, and hyperrectangles --- it excludes more expressive non-convex parameterizations. 
Moreover, each unit of the outlier budget $\rho$ of RemMCP excludes $\nq$ calibration points, and its guarantees account for $\nq$ support constraints per unit. If the calibration problem is far from fully supported (\cref{assumption:fullSupport}), i.e., fewer than $\nq$ calibration points determine the solution, RemMCP therefore becomes conservative, as observed for growing output dimensions in Appendix~\ref{subsec:varDim}. %
RelMCP supports arbitrary score functions, but introduces its own limitations: its guarantee requires a grid of penalty parameters fixed before calibration, which is difficult to choose without unnecessary conservatism, and the solution complexity $d_i$ or an upper bound thereof, which is hard to compute for non-convex problems. H-RelMCP, which we use in the experiments, avoids both requirements, but carries no proven finite-sample guarantee: its share of runs below the target coverage never exceeded $\beta$ in our experiments, yet 500 runs cannot empirically confirm small values of $\beta$. Moreover, its penalty search is not guaranteed to find a valid solution, and its conservatism depends on the initialization of the penalty parameter.

Finally, while we provide a general framework and establish its theoretical guarantees, we do not offer an extensive treatment of score and cost function design beyond a representative set of instantiations. The space of admissible score and cost functions is large, and the choice of parameterization has a significant effect on the expressiveness, computational tractability, and conservatism of the resulting prediction sets.
Importantly, MCP does not learn the geometry of the prediction sets in a model-free way: the admissible family $\Y(x,\qe)$ is determined by the chosen score function, and the cost function merely selects a preferred member of this family. Our guarantees therefore require the score function and the cost function to be fixed before calibration, e.g., based on prior knowledge or by evaluating the performance over additional data that is independent of $\Mcal$.

These limitations point to several directions for future work. On the theoretical side, a rigorous finite-sample guarantee for data-adaptive penalty selection would close the gap between RelMCP and H-RelMCP, and tighter finite-sample bounds and initialization strategies with stronger convergence guarantees would strengthen their practical reliability. On the applied side, a systematic investigation of score and cost function designs for specific output structures --- including time series, classification, and structured prediction tasks --- would substantially broaden the applicability of MCP. Beyond the two variants proposed here, other calibration schemes could be developed that may offer better trade-offs between conservatism and computational complexity. Extending MCP to settings with distribution shift is another natural and practically important direction.

\newpage

\section*{NeurIPS Paper Checklist}

\begin{enumerate}

\item {\bf Claims}
    \item[] Question: Do the main claims made in the abstract and introduction accurately reflect the paper's contributions and scope?
    \item[] Answer: \answerYes{} %
    \item[] Justification:  Please refer to the abstract and the enumeration of contributions at the end of the introduction section and compare them with the main part of the paper. %

\item {\bf Limitations}
    \item[] Question: Does the paper discuss the limitations of the work performed by the authors?
    \item[] Answer: \answerYes{} %
    \item[] Justification: Assumptions are discussed in \cref{subsec:scenario} and stated in the respective theorems that use them. Limitations are mentioned in \cref{sec:conclusions} %
    and discussed in detail in Appendix~\ref{sec:lim}. %

\item {\bf Theory assumptions and proofs}
    \item[] Question: For each theoretical result, does the paper provide the full set of assumptions and a complete (and correct) proof?
    \item[] Answer: \answerYes{} %
    \item[] Justification: All assumptions are stated in the respective theorems and the proofs are provided in Appendix~\ref{sec:proofs}. %

    \item {\bf Experimental result reproducibility}
    \item[] Question: Does the paper fully disclose all the information needed to reproduce the main experimental results of the paper to the extent that it affects the main claims and/or conclusions of the paper (regardless of whether the code and data are provided or not)?
    \item[] Answer: \answerYes{} %
    \item[] Justification: The novel methods are described in \cref{sec:multiVarCP}, further implementation details are provided in \cref{sec:experiments} and Appendix~\ref{sec:experimentalDetails}. %

\item {\bf Open access to data and code}
    \item[] Question: Does the paper provide open access to the data and code, with sufficient instructions to faithfully reproduce the main experimental results, as described in supplemental material?
    \item[] Answer: \answerYes{} %
    \item[] Justification: %
    A repository link with the complete code to reproduce the experiments is provided. %

\item {\bf Experimental setting/details}
    \item[] Question: Does the paper specify all the training and test details (e.g., data splits, hyperparameters, how they were chosen, type of optimizer) necessary to understand the results?
    \item[] Answer: \answerYes{} %
    \item[] Justification: Implementation details such as the data splits, hyperparameters, and optimizer setup are provided in \cref{sec:experiments} and Appendix~\ref{sec:experimentalDetails}.

\item {\bf Experiment statistical significance}
    \item[] Question: Does the paper report error bars suitably and correctly defined or other appropriate information about the statistical significance of the experiments?
    \item[] Answer: \answerYes{} %
    \item[] Justification: We report the results over 500 independent calibration runs as violin and box plots in \cref{sec:experiments} and Appendices~\ref{sec:addResults} and~\ref{sec:expectedResults}, and computation times as mean $\pm$ standard deviation.

\item {\bf Experiments compute resources}
    \item[] Question: For each experiment, does the paper provide sufficient information on the computer resources (type of compute workers, memory, time of execution) needed to reproduce the experiments?
    \item[] Answer: \answerYes{} %
    \item[] Justification: Information on the computer resources is provided in Appendix~\ref{sec:experimentalDetails}, and computation times are reported in \cref{sec:experiments} and Appendices~\ref{sec:addResults} and~\ref{sec:expectedResults}.
    
\item {\bf Code of ethics}
    \item[] Question: Does the research conducted in the paper conform, in every respect, with the NeurIPS Code of Ethics \url{https://neurips.cc/public/EthicsGuidelines}?
    \item[] Answer: \answerYes{} %
    \item[] Justification: The research conducted in this paper conforms, in every respect, with the NeurIPS Code of Ethics.

\item {\bf Broader impacts}
    \item[] Question: Does the paper discuss both potential positive societal impacts and negative societal impacts of the work performed?
    \item[] Answer: \answerNA{} %
    \item[] Justification: The paper does not have apparent societal impact. %
    
\item {\bf Safeguards}
    \item[] Question: Does the paper describe safeguards that have been put in place for responsible release of data or models that have a high risk for misuse (e.g., pre-trained language models, image generators, or scraped datasets)?
    \item[] Answer: \answerNA{} %
    \item[] Justification: This paper poses no such risks.

\item {\bf Licenses for existing assets}
    \item[] Question: Are the creators or original owners of assets (e.g., code, data, models), used in the paper, properly credited and are the license and terms of use explicitly mentioned and properly respected?
    \item[] Answer: \answerYes{} %
    \item[] Justification: We properly credit all external assets used in this work. In particular, we cite the original publications associated with the datasets and baseline methods used for benchmarking.

\item {\bf New assets}
    \item[] Question: Are new assets introduced in the paper well documented and is the documentation provided alongside the assets?
    \item[] Answer: \answerYes{} %
    \item[] Justification: The provided code is well documented.

\item {\bf Crowdsourcing and research with human subjects}
    \item[] Question: For crowdsourcing experiments and research with human subjects, does the paper include the full text of instructions given to participants and screenshots, if applicable, as well as details about compensation (if any)? 
    \item[] Answer: \answerNA{} %
    \item[] Justification: The paper does not involve crowdsourcing nor research with human subjects.

\item {\bf Institutional review board (IRB) approvals or equivalent for research with human subjects}
    \item[] Question: Does the paper describe potential risks incurred by study participants, whether such risks were disclosed to the subjects, and whether Institutional Review Board (IRB) approvals (or an equivalent approval/review based on the requirements of your country or institution) were obtained?
    \item[] Answer: \answerNA{} %
    \item[] Justification: The paper does not involve crowdsourcing nor research with human subjects.

\item {\bf Declaration of LLM usage}
    \item[] Question: Does the paper describe the usage of LLMs if it is an important, original, or non-standard component of the core methods in this research? Note that if the LLM is used only for writing, editing, or formatting purposes and does \emph{not} impact the core methodology, scientific rigor, or originality of the research, declaration is not required.
    \item[] Answer: \answerNA{} %
    \item[] Justification: The core method development in this research does not involve LLMs as any important, original, or non-standard components.

\end{enumerate}

\end{document}